\documentclass{article}

\usepackage[preprint]{neurips_2019}

\usepackage[utf8]{inputenc} 
\usepackage[T1]{fontenc}    
\usepackage{hyperref}       
\usepackage{url}            
\usepackage{booktabs}       
\usepackage{amsfonts}       
\usepackage{nicefrac}       
\usepackage{microtype}      
\usepackage{color}
\usepackage{float}

\usepackage{algorithmic}
 
\usepackage{hyperref}

\usepackage{wrapfig}
\usepackage{subcaption}
\usepackage{subcaption} 
\usepackage[ruled]{algorithm2e}
\usepackage{amssymb,amsmath}
\usepackage{amsthm}
\usepackage{bm}
\usepackage{dsfont}

\def\EXP{\mathbb{E}}

\usepackage{graphicx}
\graphicspath{{Plots/}}

\title{Doubly Robust Off-Policy Actor-Critic Algorithms \\
for Reinforcement Learning}

\author{
Riashat Islam\\
McGill University, Mila\\
School of Computer Science\\
\texttt{riashat.islam@mail.mcgill.ca}\\
\And
Raihan Seraj\\
McGill Univeristy\\
Electrical and Computer Engineering\\
\texttt{raihan.seraj@mail.mcgill.ca}\\
\And
Samin Yeasar Arnob\\
McGill Univeristy\\
Electrical and Computer Engineering\\
\texttt{samin.arnob@mail.mcgill.ca}\\
\And
Doina Precup\\
McGill University, Mila\\
School of Computer Science\\
\texttt{dprecup@cs.mcgill.ca}\\
}

\begin{document}

\maketitle

\begin{abstract}
We study the problem of off-policy critic evaluation in several variants of value-based off-policy actor-critic algorithms. Off-policy actor-critic algorithms require an off-policy critic evaluation step, to estimate the value of the new policy after every policy gradient update. Despite enormous success of off-policy policy gradients on control tasks, existing general methods suffer from high variance and instability, partly because the policy improvement depends on gradient of the estimated value function. In this work, we present a new way of off-policy policy evaluation in actor-critic, based on the doubly robust estimators. We extend the doubly robust estimator from off-policy policy evaluation (OPE) to actor-critic algorithms that consist of a reward estimator performance model. We find that doubly robust estimation of the critic can significantly improve performance in continuous control tasks. Furthermore, in cases where the reward function is stochastic that can lead to high variance, doubly robust critic estimation can improve performance under corrupted, stochastic reward signals, indicating its usefulness for robust and safe reinforcement learning. 
\end{abstract}

\section{Introduction}

Policy gradient based methods are widely popular in deep reinforcement learning (RL) for solving continuous control tasks \citep{trpo}. Several variants of off-policy value gradient based methods have been proposed recently \citep{SAC,lillicrap2015continuous} with the goal to solve complex manipulation while being sample efficient due to the ability to re-use off-policy data. Often deep RL policy gradient algorithms rely on using an off-policy estimate of the value function based on which the policy parameters can be directly updated by finding gradients directly through the value function. Existing literature in RL on off-policy evaluation has a long history \citep{precup2000eligibility, DR_Jiang} where the goal is to estimate the value of a policy using data sampled from another behaviour policy. Off-policy methods generally suffer from high variance due to importance sampling corrections, although several approaches have introduced bias by learning a performance model to reduce variance. Additionally, in several control and robotic tasks, the reward function may be corrupted or noisy, e.g rewards from sensors. Stochastic rewards may often make the off-policy learning process even more difficult, especially for learning complex behaviours in robotic applications. The existence of corrupted reward signals can serve as a severe bottleneck towards scaling up RL algorithms for practical applications, with the goal towards robust decision making. 

Several existing work have proposed for conservative policy iteration \citep{KakadeL02} and safe policy improvement \citep{PirottaRPC13}, where an important motivation for off-policy evaluation is to guarantee safety before the policy can be deployed in the real world world. While off-policy evaluation (OPE) approaches make use of past data for policy evaluation and have been shown to be beneficial for practical tasks, most of the success with robotic applications have come from policy gradient based methods for continuous control \citep{trpo, lillicrap2015continuous}. Therefore, it is critical to think about safety measures in off-policy actor-critic algorithms, since they have have been shown to be most sample efficient in control tasks. Several off-policy deep RL policy gradient based algorithms rely on off-policy critic evaluation, as in off-policy actor-critic based methods such as the widely used Deep Deterministic Policy Gradient (DDPG) algorithm. However, is it is surprising to see that the gap in literature between off-policy evaluation (OPE) problems and those methods being extended to the control and policy gradient setting. 

In this work, we extend the existing doubly robust estimators for off-policy evaluation (OPE) to the control setting, in off-policy actor-critic algorithms. In off-policy value-based policy gradient algorithms such as DDPG and SAC \citep{lillicrap2015continuous,SAC}, the critic evaluates the performance of a policy and the policy is improved based on the critic estimate. Often existing value-based policy gradient algorithms, however, suffer from high variance and instability particularly in continuous control tasks \citep{DRL, Islam}. This problem can be further exacerbated in practical sensory-motor driven robotic applications where the reward functions are often noisy and corrupted. As such, existing off-policy policy gradient algorithms would be quite unreliable for use in the real world. We propose doubly robust estimation for critic evaluation towards the goal of reducing variance in the critic estimates, often better stability and safe improvements in performance. Furthermore, we find that using a reward function estimator in the case of noisy rewards as demonstrated in \citep{Josh_Corl} can be quite useful for doubly robust off-policy actor-critic algorithms under stochastic rewards, further justifying that these estimators can possibly play the role of a control variate in policy gradients \citep{DR_ControlVariate}.

We aim to merge the gap between off-policy evaluation (OPE) estimators with guarantees towards unbiasedness and low variance, and estimators used in off-policy actor-critic based methods. Our goal is to achieve low variance regression based critic evaluation based on minimizing the mean squared error between next state critic target and current critic estimates, while keeping the critic estimator unbiased. We propose to use doubly robust estimators for off-policy actor-critic based methods, to show the significance of being at the intersection of model-based and model-free actor-critic algorithms. There are primarily two classes of off-policy value evaluation, where either a model-based approach is taken to fit an MDP model and evaluate the policy based on the learned model as in the \textit{Direct Method (DM)}, or approaches based on \textit{Importance Sampling (IS)} estimators which are completely model-free, unbiased and independent of the state space, but can suffer from uncontrolled variance especially in long horizon tasks. While several existing actor-critic algorithms either use model-based estimates \citep{MBPO} or use IS corrections and truncations \citep{wang2017ACER}, we propose a novel approach towards extending doubly robust estimators, based on a combination of direct model-based approach and model-free IS based estimators in the off-policy actor-critic setting for deep RL. We achieve this by proposing a reward function estimator, to estimate the reward function of the MDP, based on which we can estimate the DR estimator relying on the predicted MDP rewards. Such an approach can be particularly useful in settings where the reward function is often noisy \citep{Josh_Corl} as often found in control and robotics applications. Our key contributions are as follows: 

\begin{itemize}
    \itemsep0em 
    \item We extend the doubly robust estimators, previously proposed for contextual bandits \citep{DR_Bandits} and off-policy evaluation (OPE) in RL \citep{DR_Jiang} to the off-policy actor-critic setting in RL, for low variance critic evaluation in policy gradients
    \item We present a novel formulation for the model-based part of the reward estimator, based on using a function approximator for estimating the MDP rewards. We then derive the doubly robust (DR) estimator based on the predicted rewards and use it for minimizing the mean squared error in critic evaluation. 
    \item Our proposed doubly robust (DR) estimators for actor-critic algorithms can also be interpreted as a novel formulation as an action-dependent control variate, while keeping the policy gradient estimator unbiased  but loweing variance of the gradient estimates
    \item We find that extending the DR estimator for off-policy actor-critic can be significantly useful for reducing variance of the critic evaluation, since in most off-policy value gradient based approaches, the estimate of the critic plays a key role in performance. Our proposed extension is evaluated on a wide range of benchmark continuous control tasks, for both growing batch and fixed batch off-policy deep RL settings
    \item We find that for stochastic or corrupted reward signals, doubly robust estimators can be particularly useful, since existing methods suffer from high variance in presence of noisy reward signals. Doubly robust estimators in actor-critic can significantly reduce the variance in critic evaluation under stochastic rewards, leading to robust and safe algorithms for practical applications. We evaluate our proposed algorithm on noisy reward versions of existing control benchmark tasks, and find that using DR estimation can significantly reduce variance and improve performance. 
\end{itemize}

\section{Preliminaries}
In policy gradient methods, the aim is to learn a parameterized policy $\pi_{\theta}(a|s)$ to maximize the discounted sum of cumulative returns along the sampled trajectories, given by $J(\pi_\theta) = \EXP{\pi_\theta}[ \sum_{i=t+1}^{\infty} \gamma^{i} r(s_i, a_i)]$. Based on the policy gradient theorem \cite{sutton2000policy}, we can improve the policy parameters $\theta$ using the policy gradient, which can be computed with Monte-Carlo estimation $ \nabla_{\theta} J(\theta) = \EXP{\pi_\theta} [ \nabla_{\theta} \log \pi_{\theta}(a|s) Q^{\pi_\theta}(s,a)]$, where, the $\EXP$ above uses samples under the current policy $\pi$. Often the policy gradient estimator can suffer from high variance, and hence an advantage function $A^{\pi_\theta}(s,a) = Q^{\pi_\theta}(s,a) - V^{\pi_\theta}(s)$ or a state dependent baseline, or a combination of both is used in practice to reduce baseline $\nabla_{\theta} J(\theta) = \EXP{\pi_\theta} [ \nabla_{\theta} \log \pi_{\theta}(a|s) ( Q^{\pi_\theta}(s,a)  - V^{\pi_\theta}(s))  ] $. Alternatively, often the gradient is also estimated with an advantage function critic and using a state-action dependent baseline, such that $A^{w}(s,a) - Q(s,a)$ where the critic uses a function approximator parameterized by $w$ and a separately learned baseline is used. 

\subsection{Off-Policy Actor-Critic Algorithms}
Instead of on-policy gradient estimators, which can be sample-inefficient in practice, due to their inability to re-use data from past experiences, often an off-policy gradient estimate is preferred, based on the deterministic policy gradient (DPG) theorem \cite{Silver2014DPG}. For continuous control tasks, the DDPG algorithm \cite{lillicrap2015continuous} is often used due to their ease ability to learn from experience replay buffer. The off-policy policy gradient estimator is given by $\nabla_{\theta}J(\theta) = \EXP{\mu} [ \nabla_{\theta} Q^{w}(s, \pi_{\theta}(s)  ]$, where $Q^{w}$ is a critic estimate and $\pi_{\theta}$ is a deterministic policy which outputs continuous actions, to allow directly finding the gradient of the action-value function. Due to the instability of off-policy gradient methods with function approximators, we often require careful fine-tuning of this algorithm \cite{DRL} as the gradient estimate is directly related to the estimate of the critic. Since the DPG \cite{Silver2014DPG} algorithm can avoid importance sampling (IS) corrections, typically required in off-policy learning \cite{precup2000eligibility}, the critic can be evaluated with a regression based objective without any high variance IS corrections being required $L(w) = \EXP_{\mu} [ ( r(s,a) + \gamma Q(s', \pi_{\theta}(s')) - Q(s, \pi_{\theta}(s)) )^{2}  ]$, where the $\EXP_{\mu}$ is under samples from the experience replay buffer, and the off-policy critic evaluation is a regular one-step temporal difference (TD) based update without requiring off-policy corrections, even though we are using old samples from the replay buffer. This is due to the DPG theorem \cite{silver2014deterministic} which avoids an integral over the action space, avoiding the need for IS corrections. However, the DDPG algorithm can often be unstable to use in practice \cite{DRL}, and a state-action dependent or state dependent baseline can be used in the gradient estimate.

\subsection{Doubly Robust Off-Policy Evaluation}

In off-policy evaluation, given a fixed batch or historical data generated by some behaviour or unknown policies $\beta(a,s)$, often the goal is to produce an estimate of the value function $V^{\pi}(s)$ such that the estimator has low mean squared error (MSE) between the true value function $V^{\pi_{e}}$ and the estimated $V^{\pi}(s)$. Doubly robust estimation is an idea extended from statistics to produce regression estimates, lowering the MSE, in the case of missing or incomplete data. The idea of DR estimators extended from statistics were initially proposed for the contextual bandits setting where often the assumption was that the estimated reward function is given $\hat{R}(s,a)$, to define the DR estimator for contextual bandits $    V_{DR} = \hat{V}(s) + \rho \bigr[  r - \hat{R}(s,a)   \bigl]$, where $\hat{R}(s,a)$ is the estimated reward, $\rho$ being the IS corrections for the mismatch in the action distributions. From there, DR estimators were extended to the off-policy evaluation in RL setting \cite{DR_Jiang},\cite{DR_Phil} to reduce the variance of off-policy evaluation, while keeping the regression based estimators unbiased. \cite{DR_Jiang} argued that instead of using importance sampling corrections, which is unbiased, but can have high variance, it is better to use DR estimators in off-policy evaluation tasks. The key step in DR estimator is to use the following unbiased estimator 

\begin{equation}
\label{eq:original_dr_update}
V_{DR}(s) = \hat{V}(s) + \rho [ r(s,a) + \gamma V_{DR}(s') -  \hat{Q}(s,a) ]
\end{equation}

where we replace $V^{\pi}$ with $V_{DR}^{\pi}$ to denote a DR estimation of the off-policy evaluation. A key requirement in DR estimators is to use an approximation to the MDP model since the $\hat{V}$ requires the rewards from an approximation of the MDP. In other words, $\hat{R}$ used to compute $\hat{V}$ is the model's prediction of the reward. Given the samples from past data and an approximate model of the MDP, the goal of DR estimators is to produce a low variance regression mean equated error estimate $MSE(V_{DR}, V^{\pi})$. 


\section{Related Work}
Off policy evaluation in Markov decision processes is the task of evaluating a expected return of one policy with data generated by a different \textit{behavior policy}. In~\cite{Hanna}, the authors propose a regression importance sampling method where the behavior policy is estimated using the same set of data which are used for calculating the importance sampling estimate. Such an estimate of the behavior policy leads to a lower mean squared error for off policy evaluation compared with the true behavior policy. Methods related to regression importance sampling had been studied for Monte Carlo methods in~\cite{henmi2007importance,delyon2016integral}. Doubly Robust estimators for off policy value evaluation had been studied in~\cite{DR_Jiang}, where the authors used Doubly Robust estimates for Bandits as a control variate for variance reduction. An extension of this work was proposed in ~\cite{DR_Phil}, which uses doubly robust estimator and proposes a way to mix between model based estimates and importance sampling based estimates to predict the performance of a policy with historical data where the data was generated using a behavior policy. Since Doubly Robust estimators, require a reward predictor, we have adapted the reward estimator method in~\cite{Josh_Corl}. Other data driven approaches in reward estimation has been discussed in~\cite{fu2018variational,hadfield2017inverse,sermanet2016unsupervised}. A more extensive view of the Doubly Robust estimator has been proposed by~\cite{farajtabar2018more} where the authors present the formulation for learning DR model in RL and the model parameters are learned by minimizing the variance of the DR estimators. 

\section{Approach}
In this work, we extend the existing value-based off-policy policy gradient algorithms such as DDPG \citep{lillicrap2015continuous} and Soft Actor-Critic (SAC) \citep{SAC} with a doubly robust (DR) estimation of the critic $Q_{\phi}^{DR}$. We propose to use DR based critic estimation in the critic evaluation step, to reduce the variance of the critic estimate in value-based policy gradient algorithms. Since value-based gradient algorithms relies on directly finding the gradient of the action-value function, we hypothesize that reducing the variance of critic estimation can significantly improve the performance and lead to better stability in these algorithms. 

\subsection{Policy Gradient with Doubly Robust Estimator}

In this section, we derive the doubly robust estimator for policy gradient algorithms. The key idea of this estimator comes from observing equation \ref{eq:original_dr_update} of doubly robust estiimation in OPE which provides an unbiased but low variance estimation of the value function. This depends on the estimated reward function $\hat{R}(s,a)$, where accurate estimation of the MDP rewards can reduce the variance of the value function. In the next section, we will discuss our approach to estimate the MDP rewards $\hat{R}(s,a)$ for a practical algorithm. By using the unbiased DR estimator as a control variate \citep{DR_ControlVariate}, for the off-policy actor-critic algorithm, we can achieve a low variance  unbiased estimator of the critic, which helps to improve the stability of existing off-policy policy gradient algorithms. The policy gradient update is given by

\begin{equation}
\small  \nabla_{\theta} J(\theta) = \mathbb{E}_{s \sim d_{\beta}} \Big[ \nabla_{\theta} Q_{\phi}^{DR}(s, \pi_{\theta}(s) \Big]    
\end{equation}

where the critic and policies are separately parameterized with $\phi$ and $\theta$, and we update the policy parameters with stochastic gradient optimization. Considering algorithms such as DDPG, here we denote the policy improvement phase with deterministic policies $\pi_{\theta}(s)$, which has been extended to stochastic Gaussian policies with a reparameterization trick in algorithms such as Soft Actor-Critic (SAC) \citep{SAC}. The key step in our algorithm is that, we replace the critic as in DDPG with a doubly robust estimation of the critic denoted by $Q_{\phi}^{DR}(s, \pi_{\theta}(s))$, such that the critic now mininizes the mean squared regression loss with the following TD error :

\small\begin{equation}
\label{eq:critic_dr_update}
\small Q_{\phi}^{DR} (s,a) =\hat{Q}(s, \pi_{\theta}(s)) + \Bigl[ r(s,a) + \gamma Q_{\phi}^{DR}(s', \pi_{\theta}(s')) - \hat{V}(s) \Bigr]
\end{equation}

where $\hat{Q}(s, \pi_{\theta}(s))$ is following the DR estimate for action-value functions $\hat{Q}$ instead of the value function $\hat{V}$. Equation \ref{eq:critic_dr_update} shows that the critic update for the policy gradient now requires a separate estimation action-value $\hat{Q}$ and value function $\hat{V}$ as well, based on the predicted MDP rewards \cite{DR_Jiang}, ie, model-based estimation of the reward function, for the doubly robust estimation of the critic. In the next section, we discuss our approach for approximation of the rewards $\hat{R}(s,a)$. 

\subsection{Reward Function Approximator}

Since the doubly robust estimation of the critic requires an approximate MDP model, we estimate the true rewards of the mDP $R(s,a)$ with an approximate reward $\hat{R}(s,a)$ by using a separately parameterized function approximator with parameters $\psi_{R}$,  denoted by $\hat{R} = f_{\psi_{R}}(s,a)$. The MDP rewards are estimated based on samples from the replay buffer, and we use a similar approach as in \citep{Josh_Corl} where we train the reward function estimator based on the following regression loss

\begin{equation}
\label{eq:learn_R_hat}
 \small  L(\psi_{R}) = \mathbb{E}_{s, a, r \sim Buffer} [ (\hat{R}(s,a) - R(s,a))^2]
\end{equation}

We then use this reward for further estimating the approximated action-value function $\hat{Q}(s,a)$ and value function $\hat{V}(s)$ that are required for the DR estimation. Note that, to estimate this, which is a form of the advantage function or a control variate, we typically use a separate function approximator for the control variate estimate. We use another separately parameterized network with parameterss $\psi_{QV}$ which outputs both the approximated $\hat{Q}$ and $\hat{V}$ and trained with the samples from the replay buffer

\begin{equation}
\label{eq:learn_Q_hat}
   \small \mathcal{L_{\psi_{QV}}} = \mathbb{E}_{s,a,r,s' \sim Buffer}[ ( \hat{R}(s,a) + \gamma \hat{Q}(s', \pi_{\theta}(s')) - \hat{Q}(s,a)  )^{2}  ]
\end{equation}

where we use the approximated reward $\hat{R}(s,a)$ in the TD error for miniziming the loss $\mathcal{L}(\psi_{QV})$. Based on minimizing the losses in equation \ref{eq:learn_R_hat} and \ref{eq:learn_Q_hat}, we therefore get an approximation of $\hat{R}$, $\hat{Q}$ and $\hat{V}$ that are required for the doubly robust critic estimation in equation \ref{eq:critic_dr_update}.

\subsection{Algorithm}

Our full algorithm is as given below. Since our entire algorithm requires actor and critic updates, additional estimates of$\hat{R}$, $\hat{Q}$ and $\hat{V}$, we are effectively introducing a three time-scale algorithm. Therefore, in our algorithm, to ensure convergence, we use the large learning rate $\alpha_{\hat{Q}}$ for the approximated $\hat{Q}$, followed by a larger learning rate for the critic estimate compared to the actor as in actor-critic algorithms. In our algorithm, as detailed below, we therefore require the MDP rewards $\hat{R}(s,a)$ to be predicted first, based on which we can train the function approximator with a TD error based on $\hat{R}(s,a)$ to compute $\hat{Q}(s,a)$ and $\hat{V}(s)$. We then use the estimates from the model for off-policy DR evaluation of the critic $Q_{\phi}^{DR}$ such as to get unbiased but low variance estimates of the critic. Following this, we can then use \textit{any} off-policy value-based policy gradient algorithm including DDPG, SAC or TD3. 

\textbf{Requirement of Independence : } Following \citep{DR_Jiang}, note that, we use a different set of samples from the replay buffer to estimate $\hat{R}(s,a)$, $\hat{Q}(s,a)$ and $\hat{V}(s)$ than the samples used for actor and critic updates. Although this is not a strict requirement that the samples used for estimating $\pi_{\theta}(a,s)$ and $\hat{R}(s,a)$ be independent of each other, we find that using separate random samples for the model-based estimates and the model-free updates typically works better in practice. 

\begin{algorithm}[!htb]
\caption{Off-Policy Actor-Critic Algorithms with Doubly Robust Critic Estimator }
\begin{algorithmic}
\label{algo:training}
\REQUIRE ~~ A policy network $\pi_{\theta}(a,s)$, critic network $Q_{\phi}(s,a)$, and networks $\hat{R}_{\psi_{R}}$ and $\hat{Q}_{\psi_{QV}}(s,a)$
\REQUIRE ~~ The number of episodes, $E$ and update interval, $N$.
\FOR{$e=1$ to $E$}
\STATE{Take action $a_{t}$,get reward $r_{t}$ and observe next state $s_{t+1}$}
\STATE{Store tuple ($s_t,a_t, r_{t+1}, s_{t+1}$) as trajectory rollouts or in replay buffer $\mathcal{B}$}\\
{Estimate the MDP rewards $\hat{R}$, by minimizing loss \ref{eq:learn_R_hat}}\\
{Estimate approximate $\hat{Q}$ and $\hat{V}$, minimize loss \ref{eq:learn_Q_hat}}\\
{Estimate critic $Q_{\phi}^{DR}(s,a)$ using $\hat{Q}$ and $\hat{V}$ for DR estimation, by minimizing loss \ref{eq:critic_dr_update} }\\
{Update policy parameters $\theta$ following \textit{any} value-based policy gradient method such as DDPG, SAC, TD3 according to $\nabla_{\theta} \tilde{J}(\theta) = \Bigl[  \nabla_{\theta} Q_{\phi}^{DR}(s, \pi_{\theta}(s))   \Bigr]$}
\ENDFOR 
\end{algorithmic}
\end{algorithm}

\subsection{Stochastic and Noisy Rewards}

We further consider the case of using stochastic and noisy rewards, where conventional algorithms often fail due to high variance estimates of the gradient. We consider the setting where in addition to the MDP rewards $R(s,a)$, we add a Gaussian noise $\mathcal{N}(\mu, \sigma)$ to the true rewards, to make a corrupted version of the rewards. This is similar to the Corruped Reward MDP (CRMDP) \citep{corrupted_rewards}, similar to the stochastic reward control experiments considered in \citep{Josh_Corl}. This is similar to many practical robotic tasks where the reward function may often be noisy due to noise in the sensory data. We further examine the significance of using the DR estimator in cases with noisy rewards, given by $\tilde{r}(s,a) = r(s,a) + \mathcal{N}(\mu, \sigma)$ and examine the significance of DR estimator to reduce the variance of noisy critic estimates. For our experiments, we add noisy rewards to all the benchmark control Mujoco tasks, and examine how standard off-policy algorithms such as DDPG and SAC perform in presence of stochastic rewards. All our experiments, as discussed below are for noisy rewards with $\sigma=0.5$ and $\sigma=1.0$.

\section{Experimental Results}
\label{sec:result}
In all our experiments, we compare with existing value-based off-policy gradient algorithms including DDPG, SAC and TD3, and highlight the significance of reducing variance of the critic estimate with DR estimator, in terms of performance improvement. We evaluate the performance of different actor critic algorithms with a doubly robust critic estimator on several continuous control Mujoco tasks \cite{todorov2012mujoco}. Experiments are evaluated on the 
Half-Cheetah-v1, Walker-2d-v1, Hopper-v1, environments, and evaluated an average over 3 runs with random seeds.

\begin{figure}[hbt!]
\centering
   \includegraphics[width=0.3\linewidth]{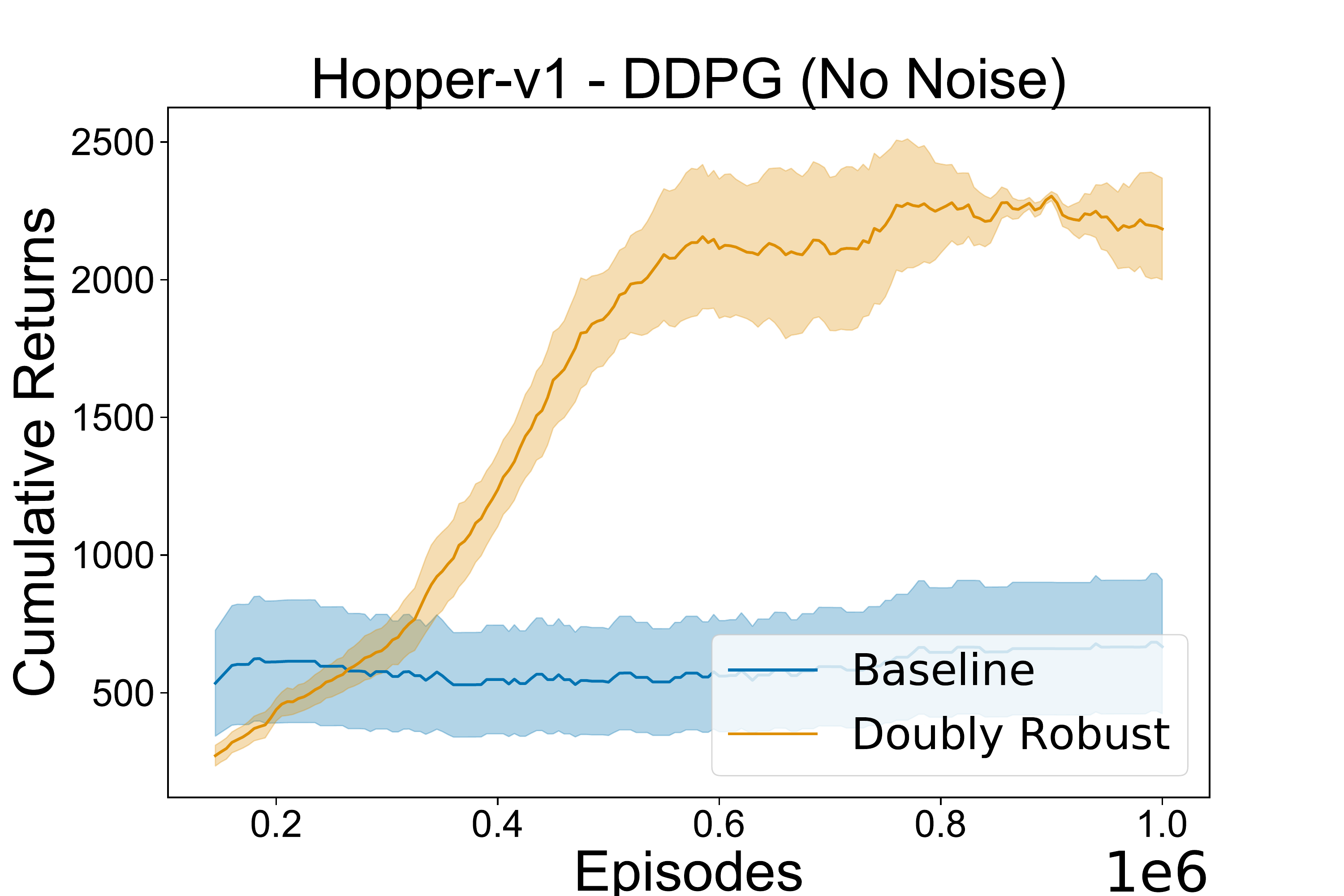}
   \!
   \includegraphics[width=0.3\linewidth]{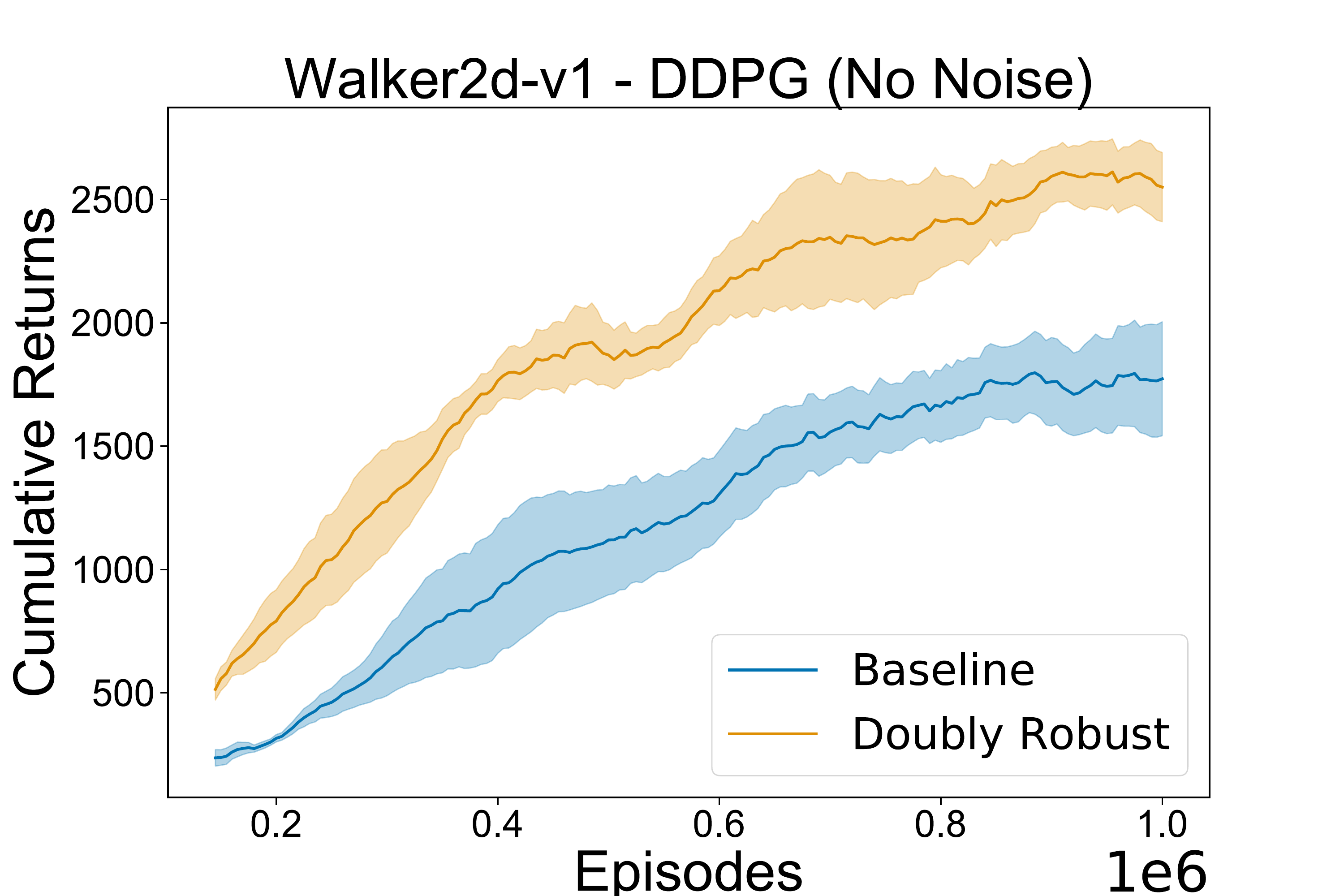}
   \!
   \includegraphics[width=0.3\linewidth]{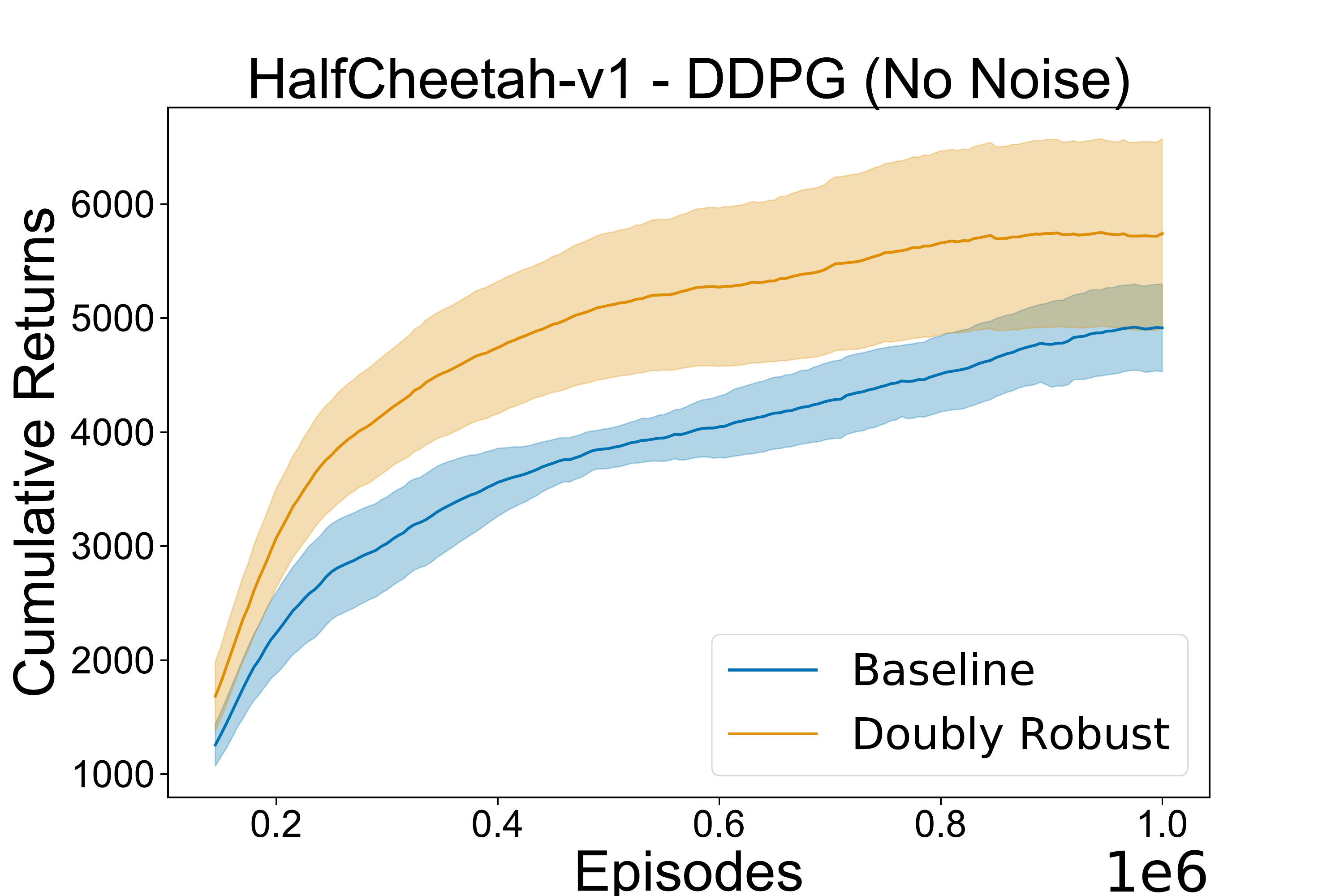}
    \!
   \includegraphics[width=0.3\linewidth]{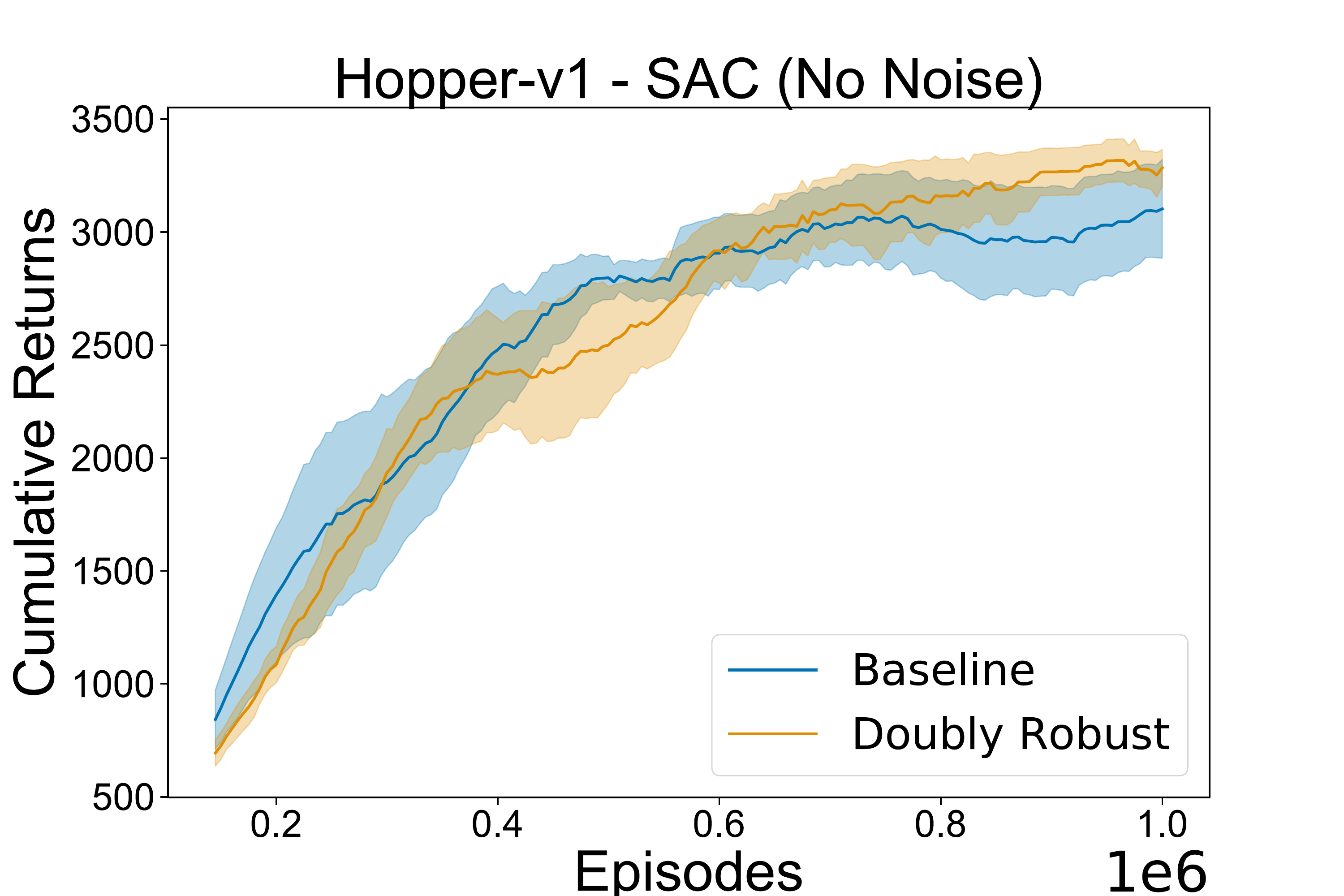}
   \!
   \includegraphics[width=0.3\linewidth]{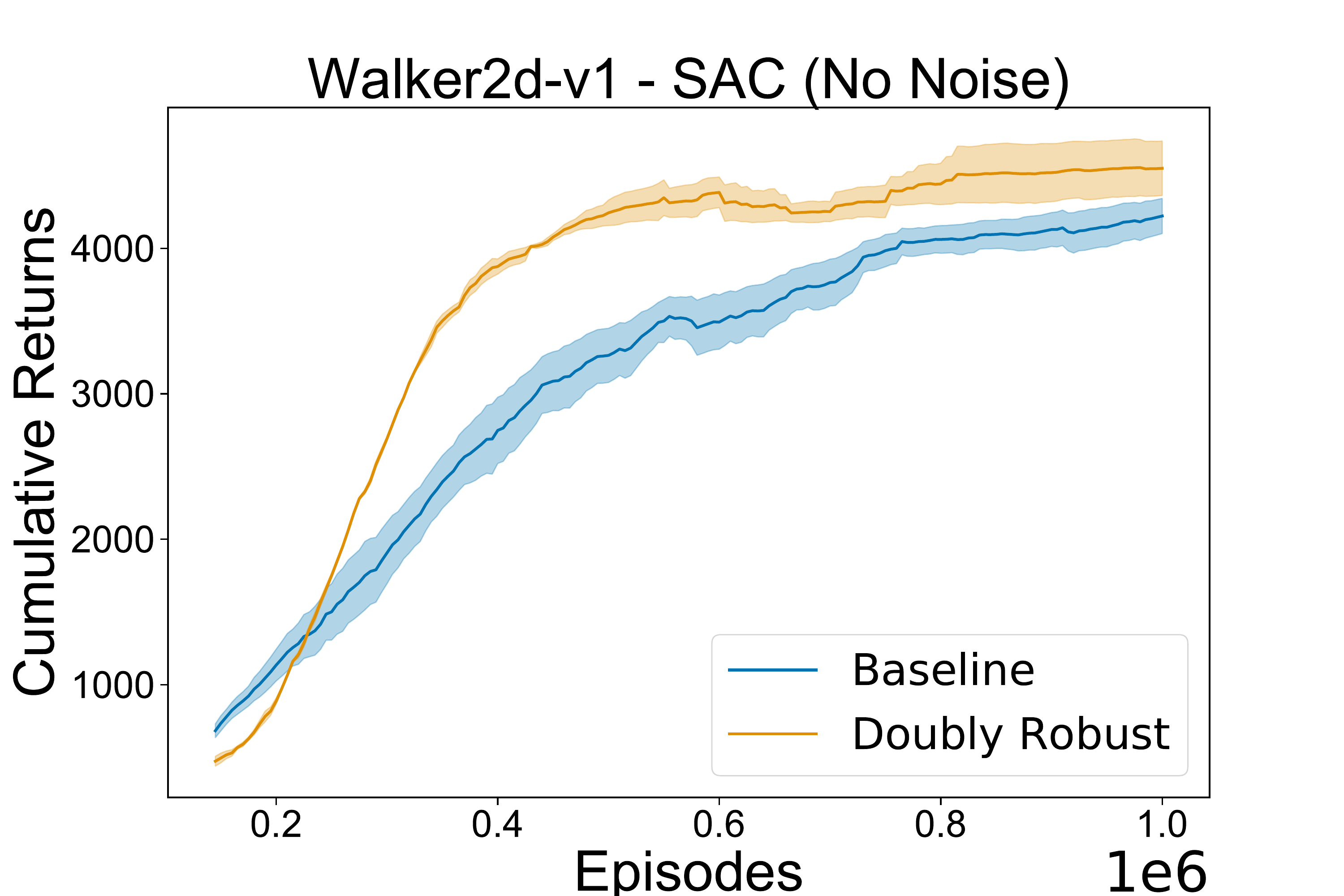}
   \!
   \includegraphics[width=0.3\linewidth]{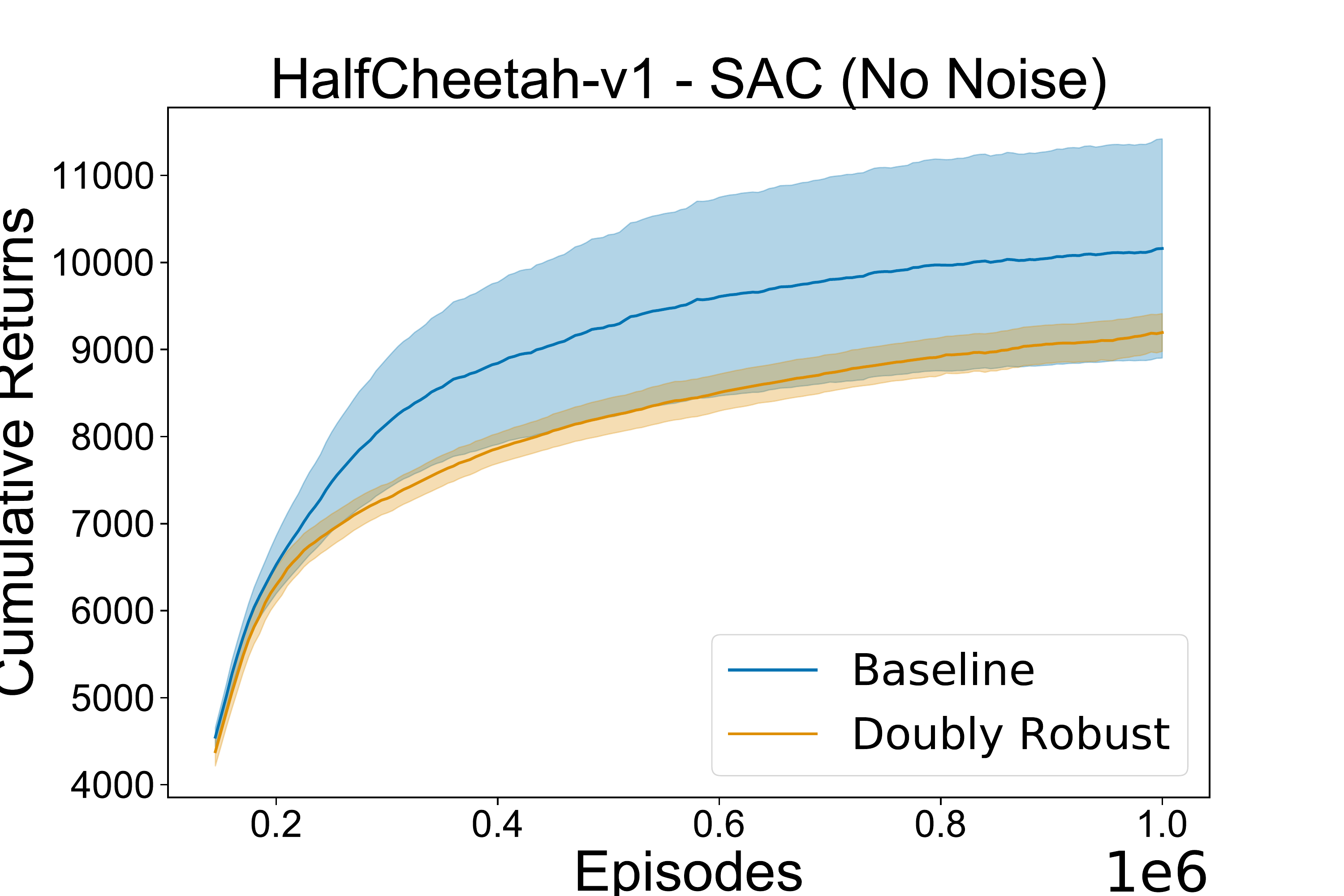}
   \!
   \includegraphics[width=0.3\linewidth]{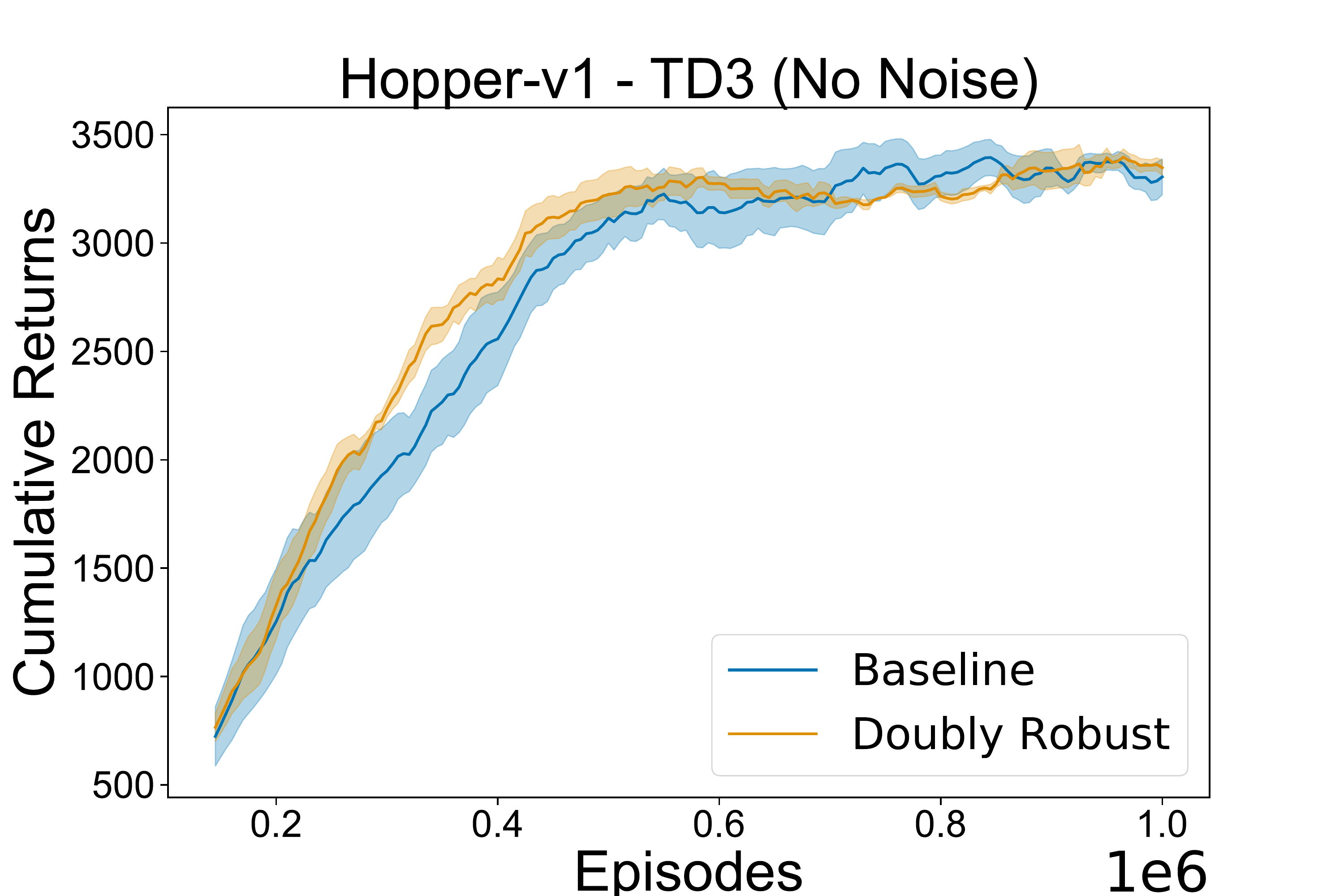}
     \!
   \includegraphics[width=0.3\linewidth]{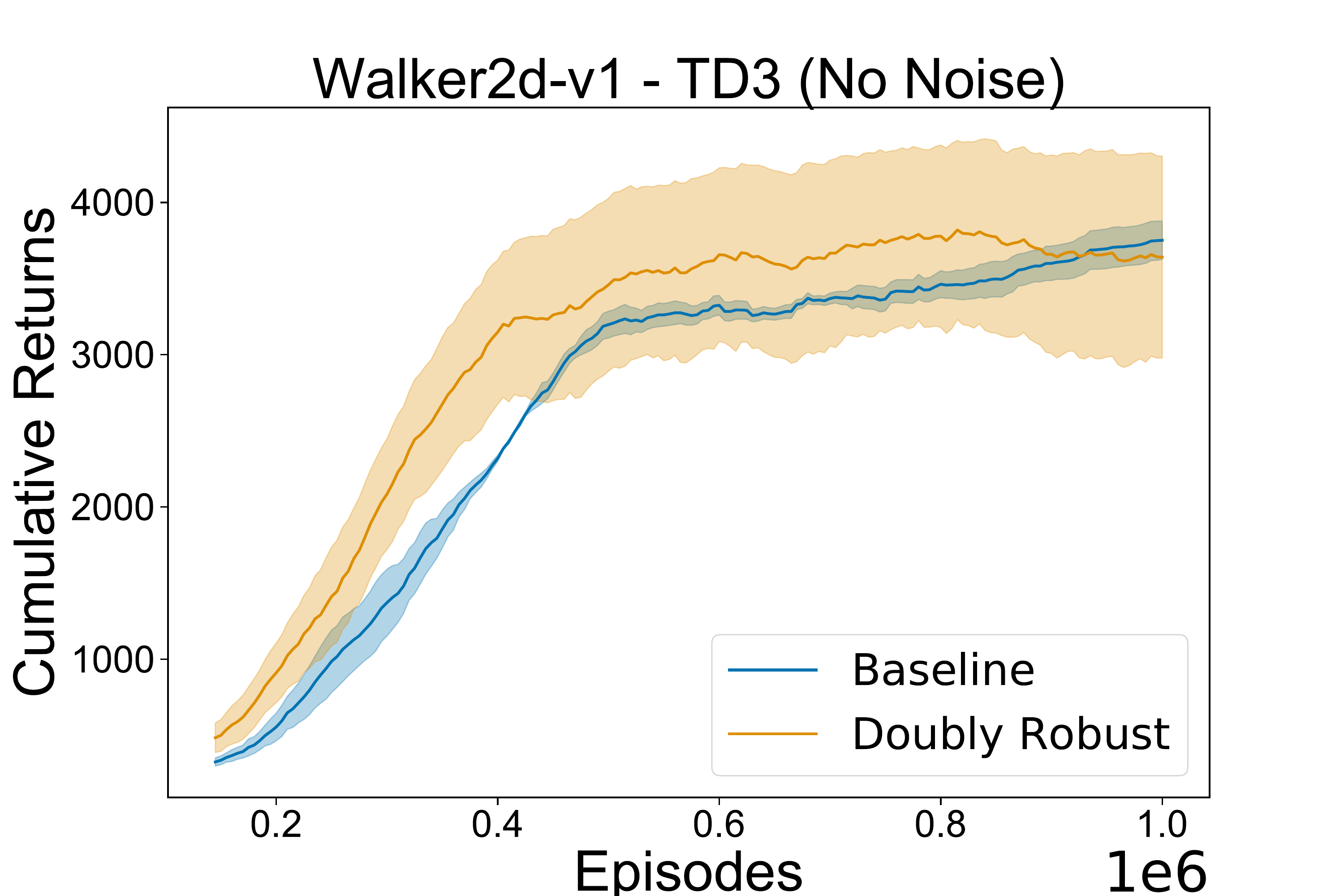}
   \!
    \includegraphics[width=0.3\linewidth]{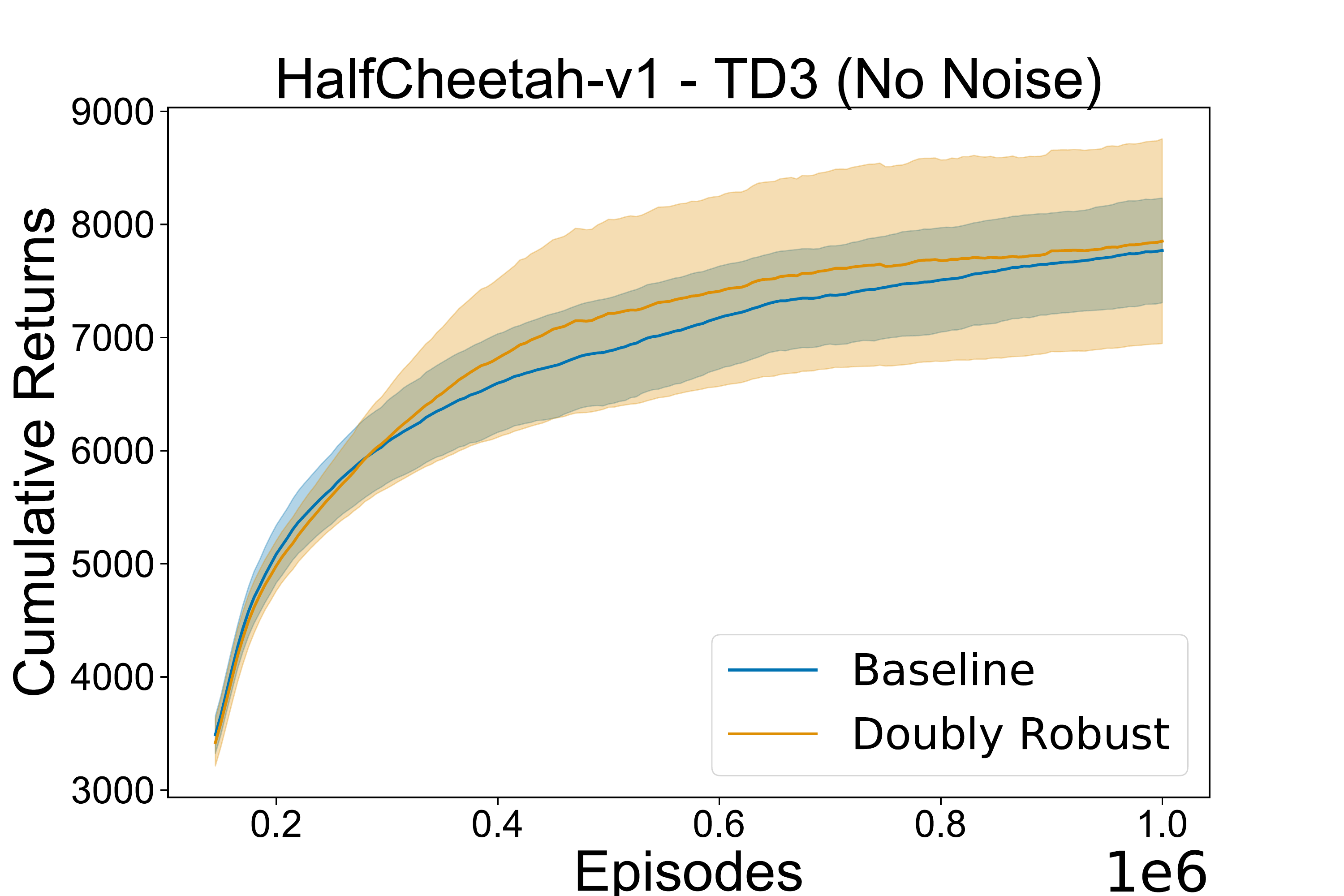}
\caption{Performance comparison of DDPG,SAC and TD3 with DR estimator using rewards with no noise}
\label{no_noisy_rewards}
\end{figure}

Figure \ref{no_noisy_rewards} shows that using the DR estimator, we can obtain improvements in the performance over standard baselines. In case of SAC and TD3 algorithms, for Hopper-v1, we obtain an equivalent performance to that of the baselines. For experiments with the HalfCheetah-v1, environemnts, we observe that the baselines for SAC is higher than our DR estimator. These results show the case where the per step rewards are exactly observed by the agent (with out any noise) and that the agent directly use these rewards in the reward estimator. We also observe that for most of the mujoco environments, the variance of the DR estimator is lower compared to the baselines. 

\begin{figure}[hbt!]
\centering
   \includegraphics[width=0.3\linewidth]{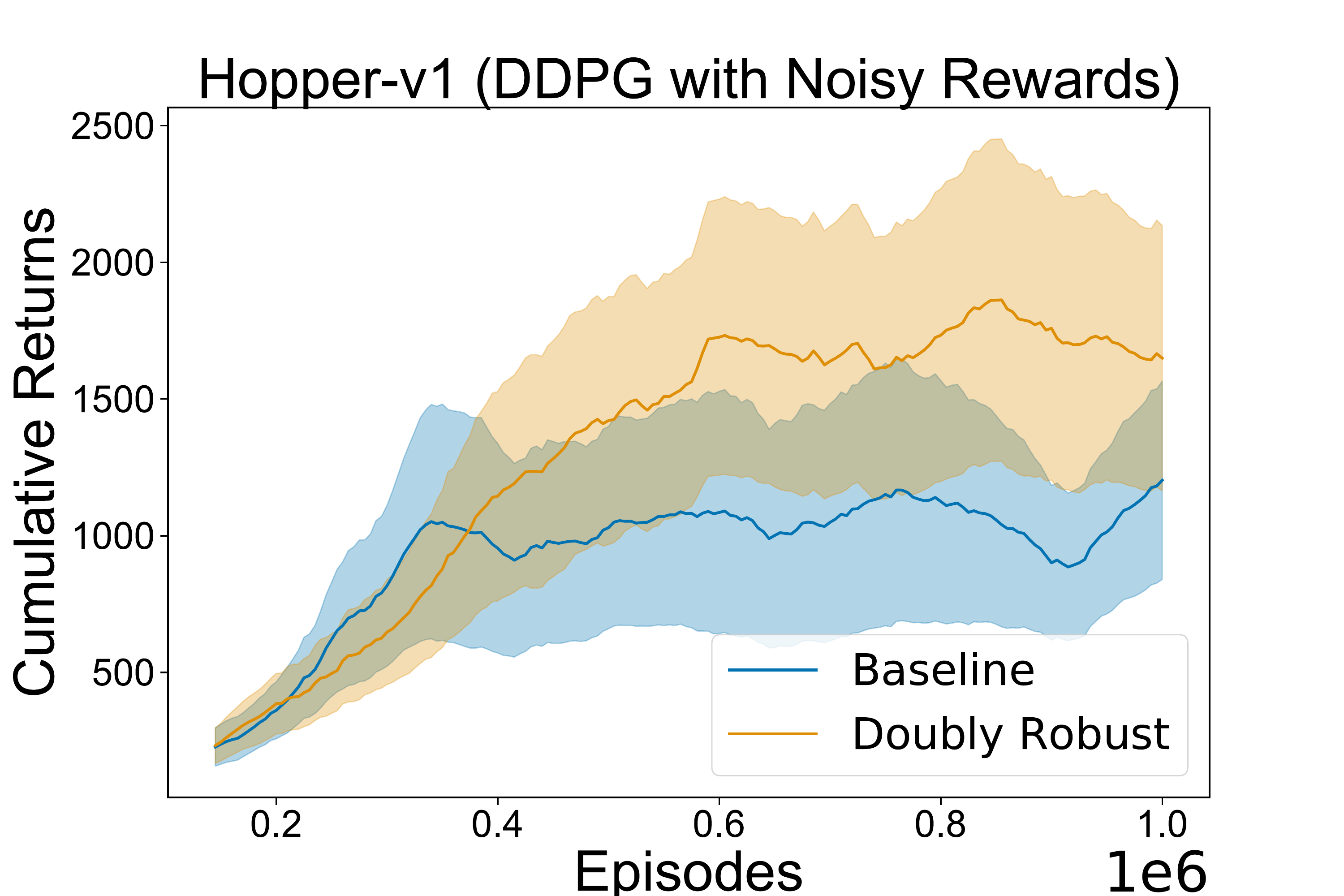}
   \!
   \includegraphics[width=0.3\linewidth]{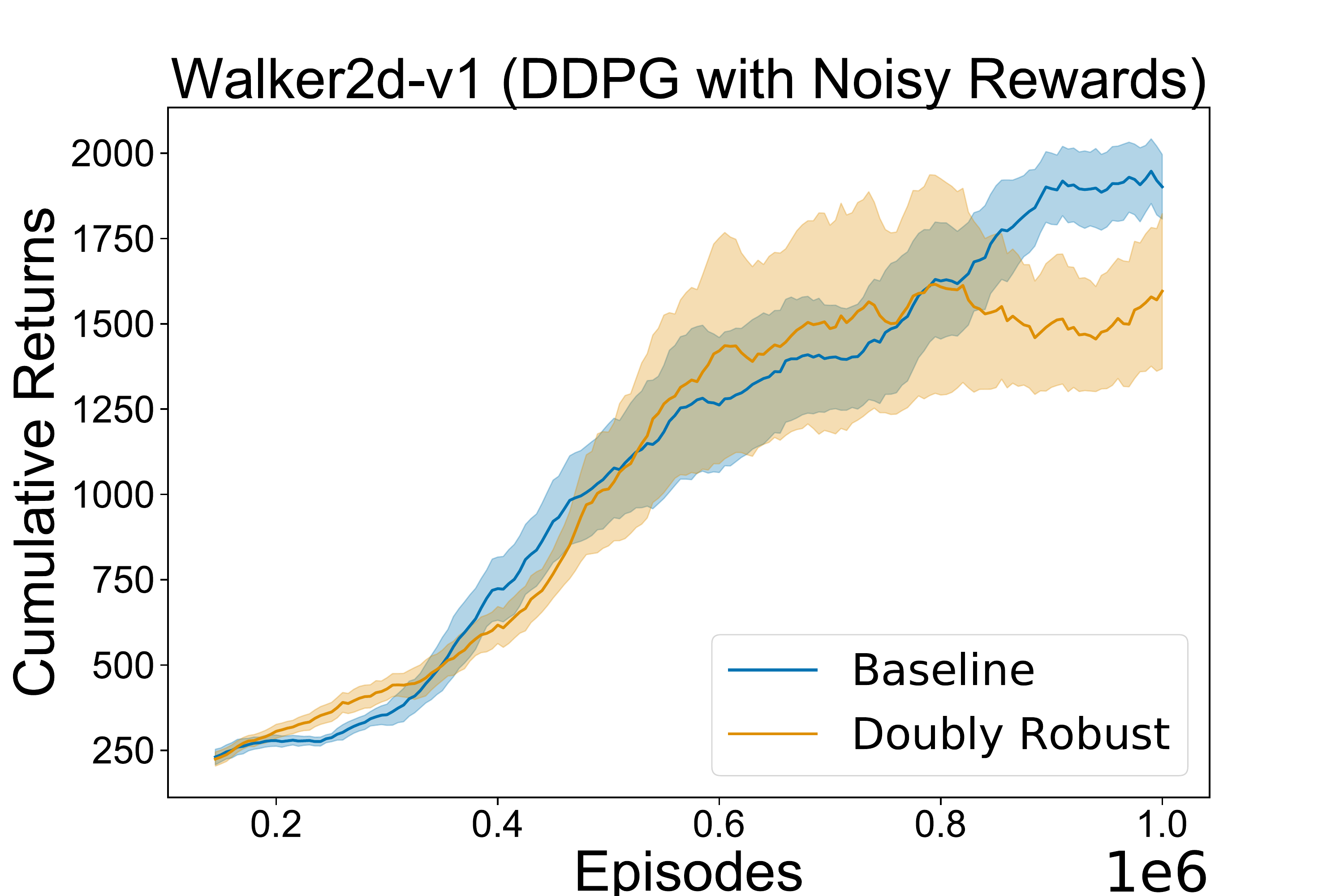}
   \!
   \includegraphics[width=0.3\linewidth]{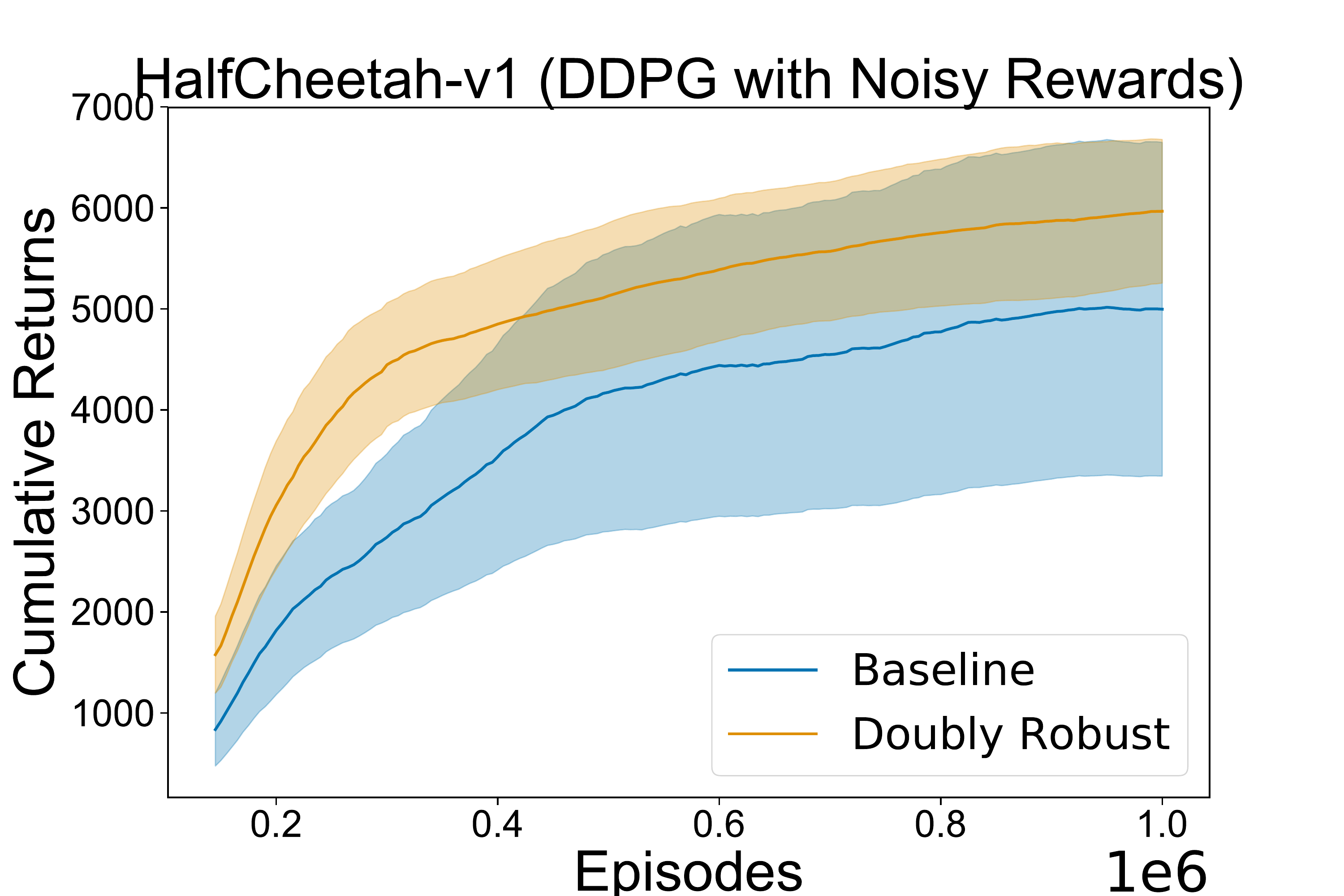}
    \!
   \includegraphics[width=0.3\linewidth]{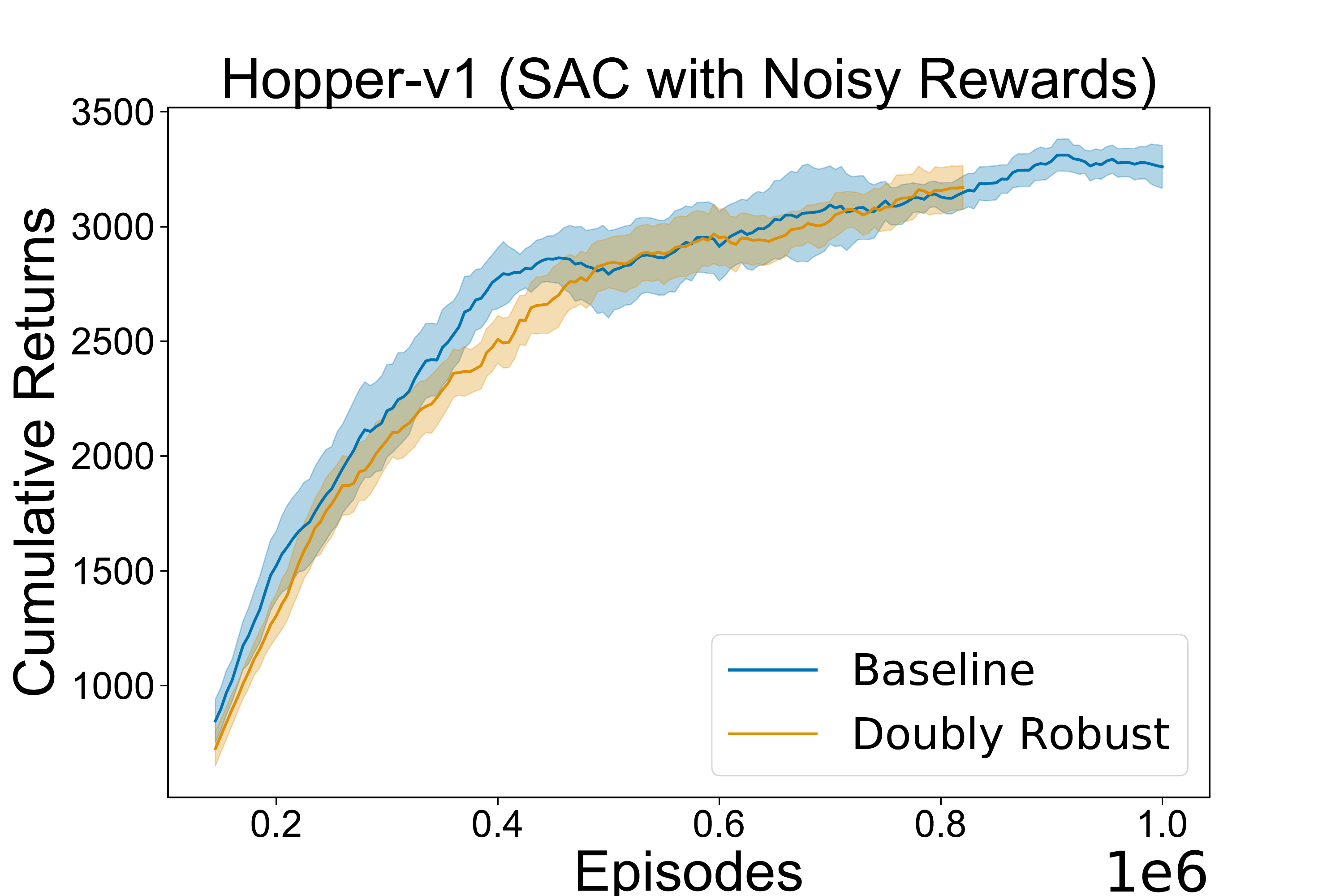}
   \!
   \includegraphics[width=0.3\linewidth]{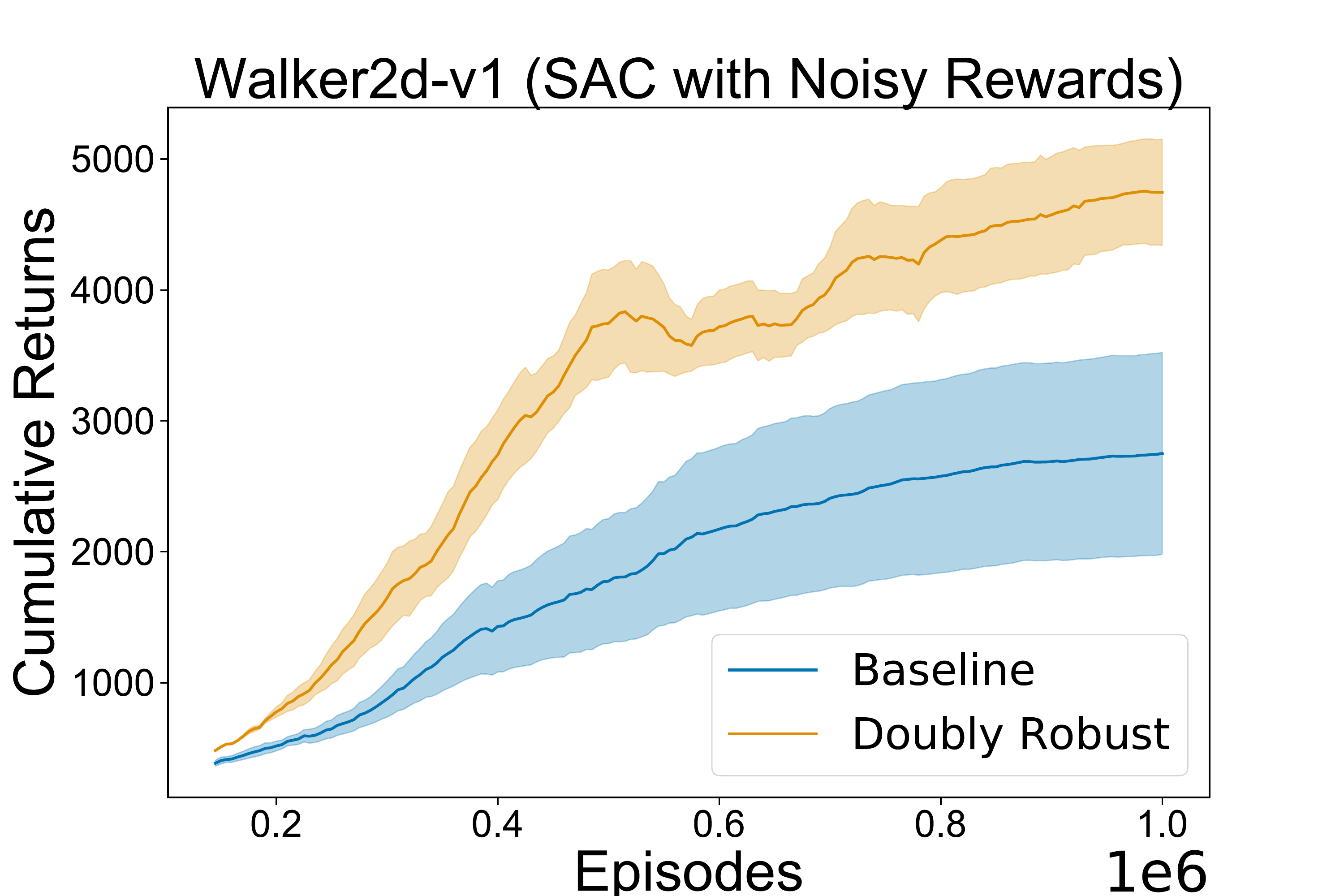}
   \!
   \includegraphics[width=0.3\linewidth]{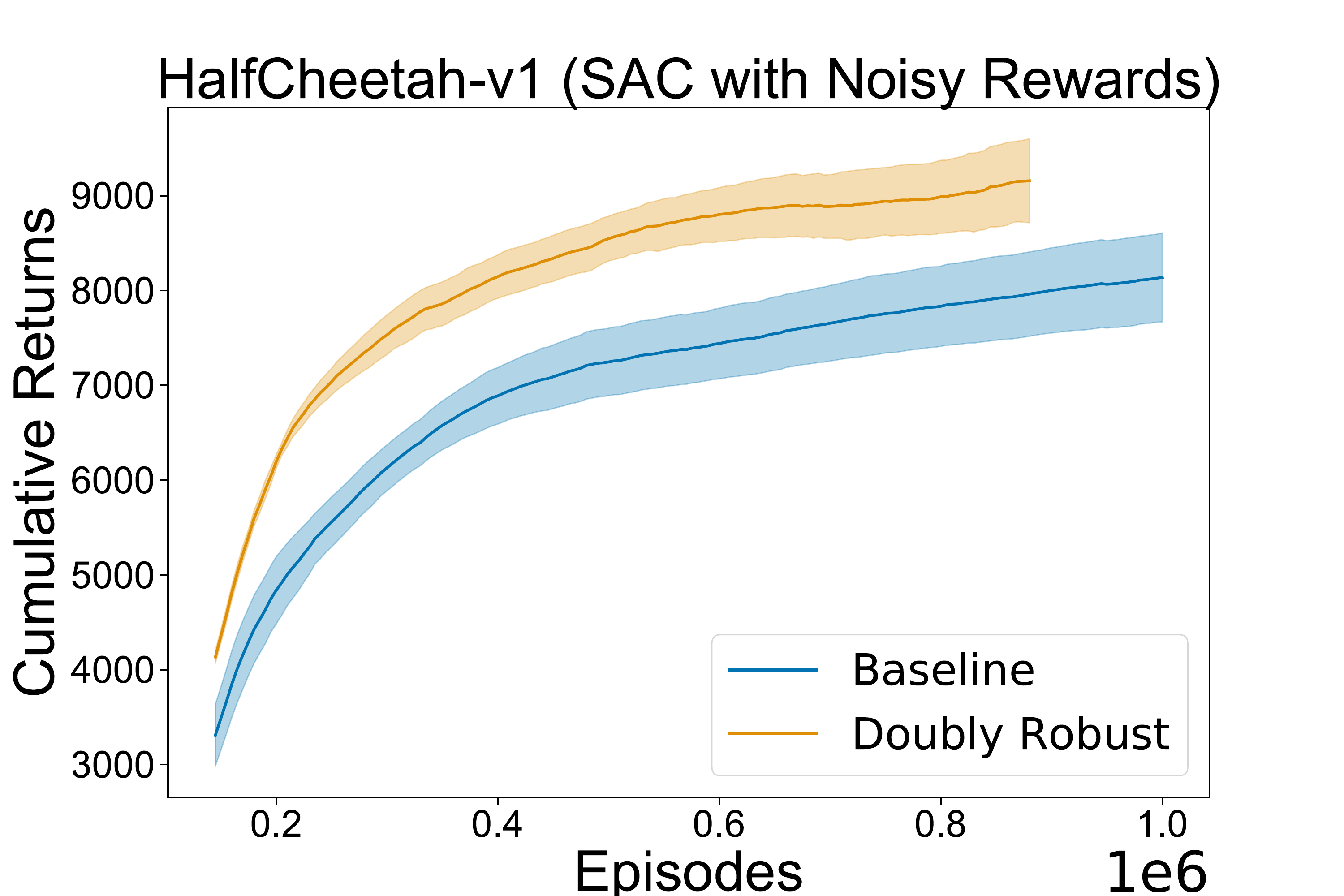}
   \!
   \includegraphics[width=0.3\linewidth]{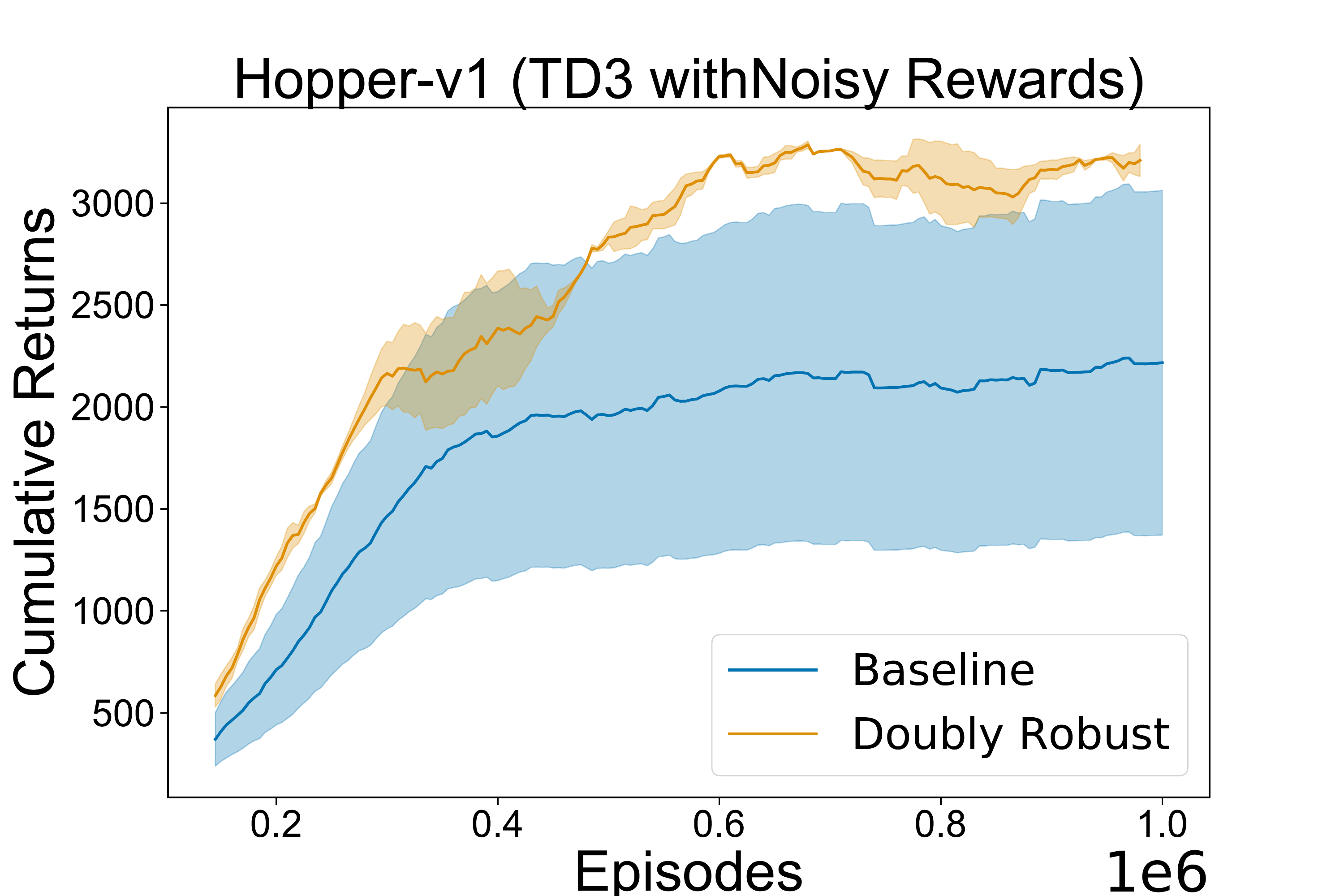}
     \!
   \includegraphics[width=0.3\linewidth]{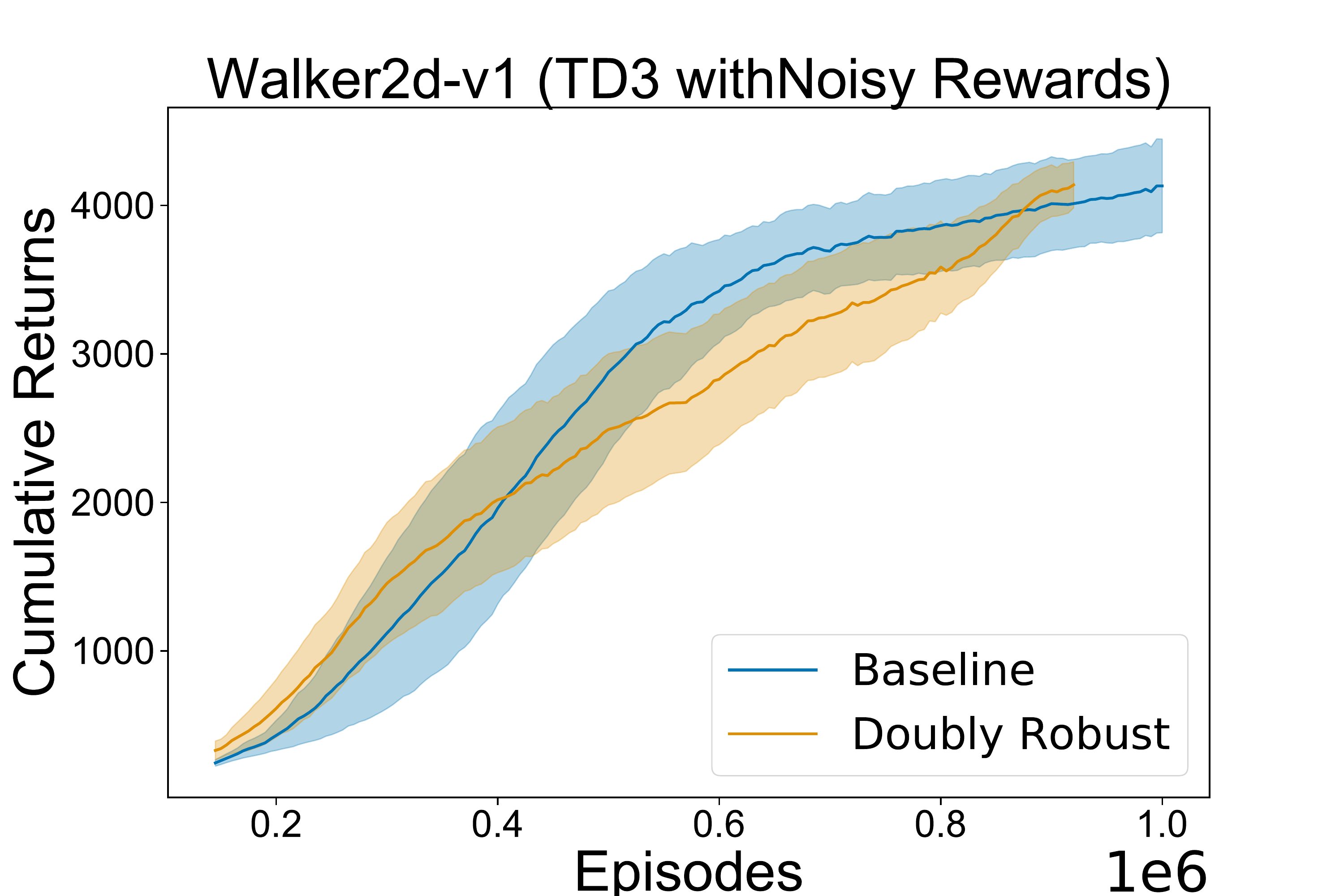}
   \!
    \includegraphics[width=0.3\linewidth]{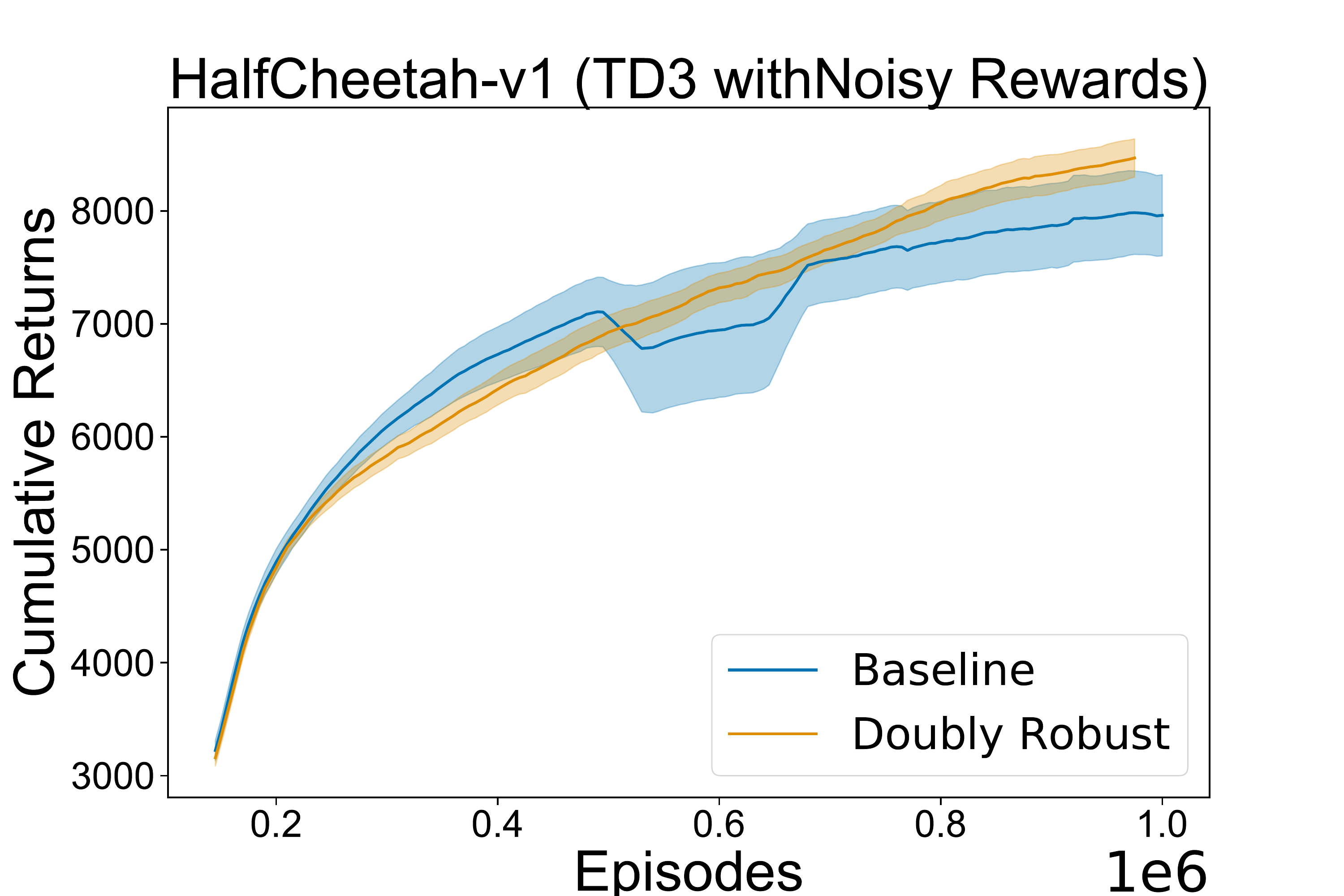}
\caption{Performance comparison of DDPG, SAC and TD3 with DR estimator using noisy rewards,with $\sigma=0.5$}
\label{low_noisy_rewards}
\end{figure}

\begin{figure}[hbt!]
\centering
   \includegraphics[width=0.3\linewidth]{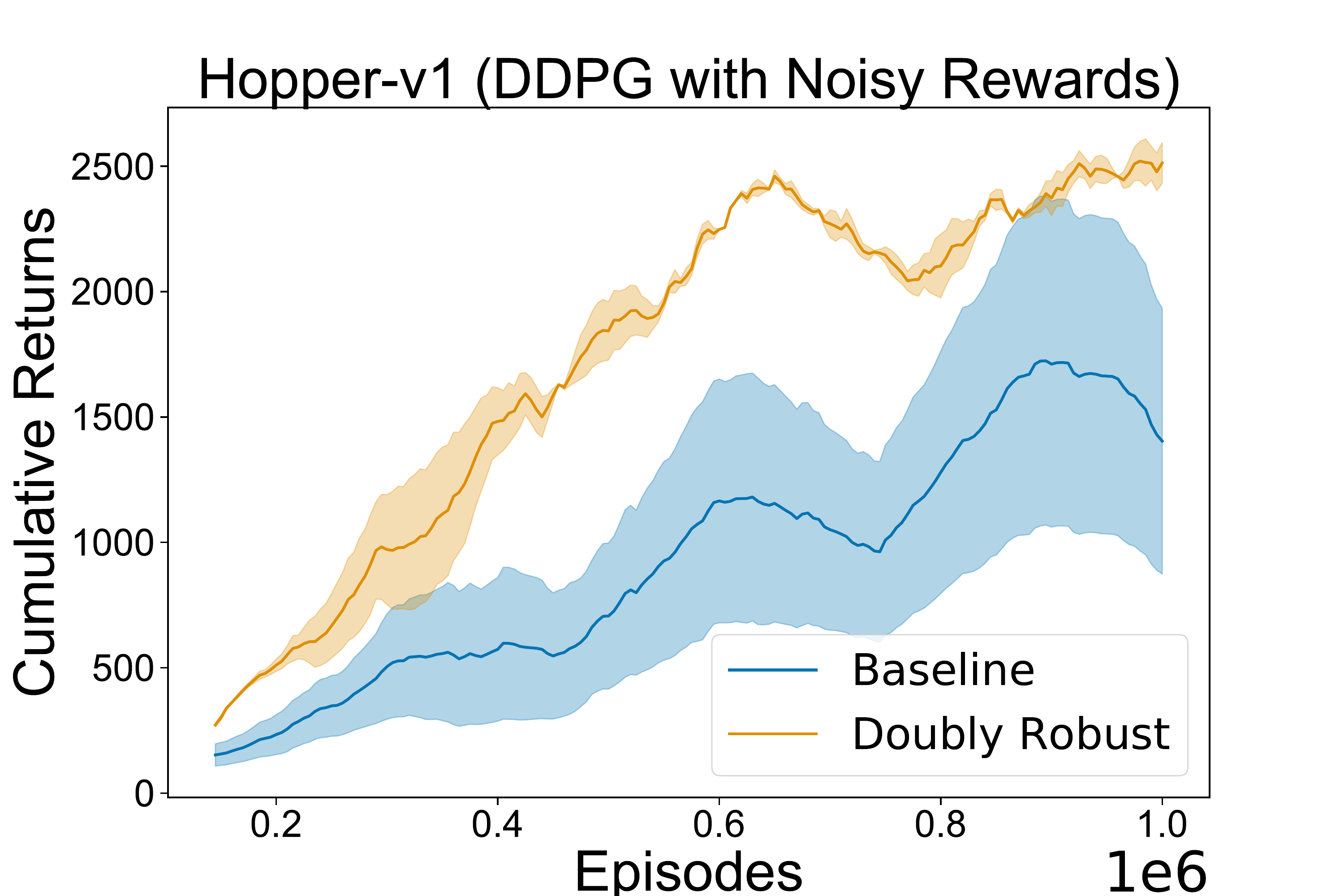}
   \!
   \includegraphics[width=0.3\linewidth]{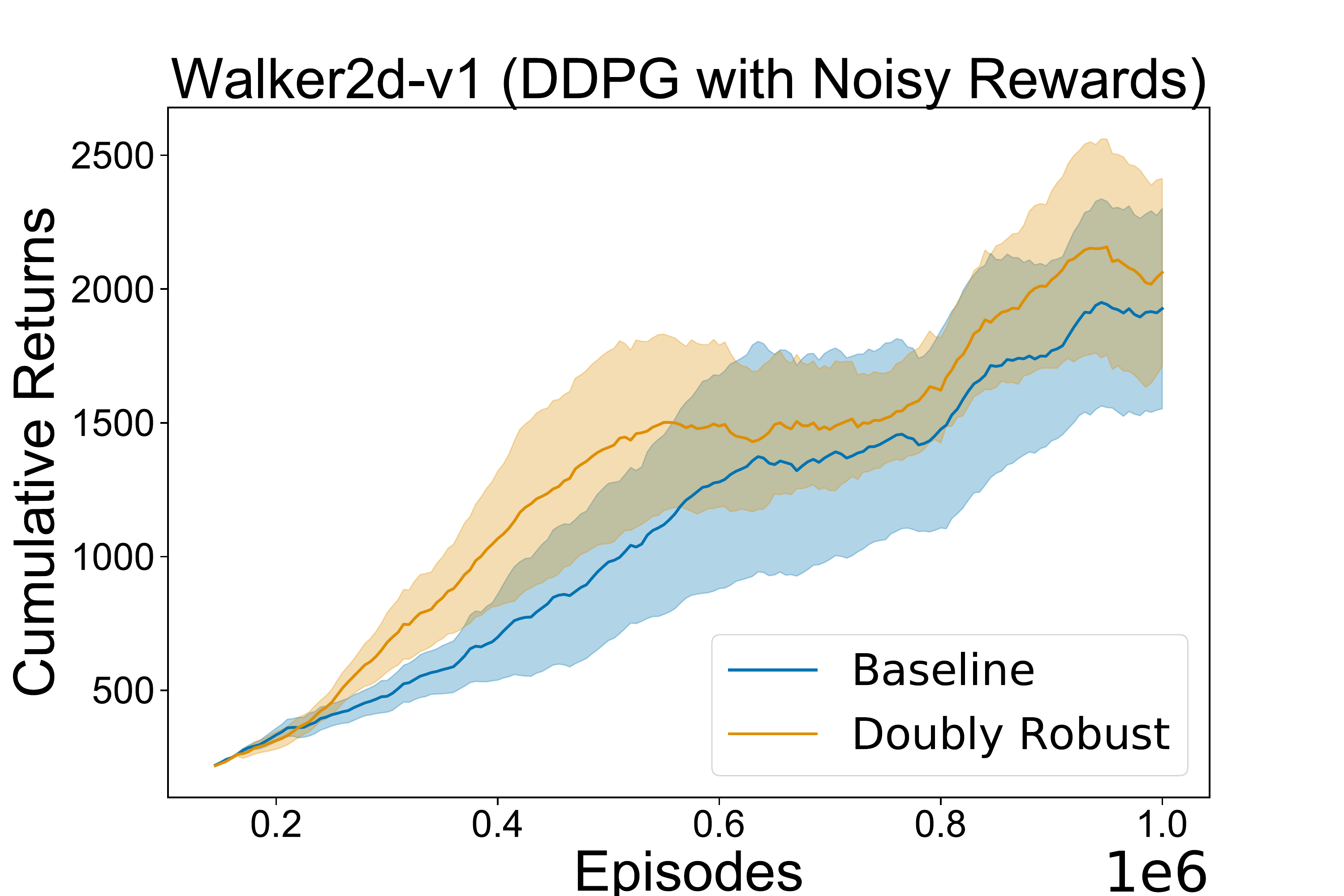}
   \!
   \includegraphics[width=0.3\linewidth]{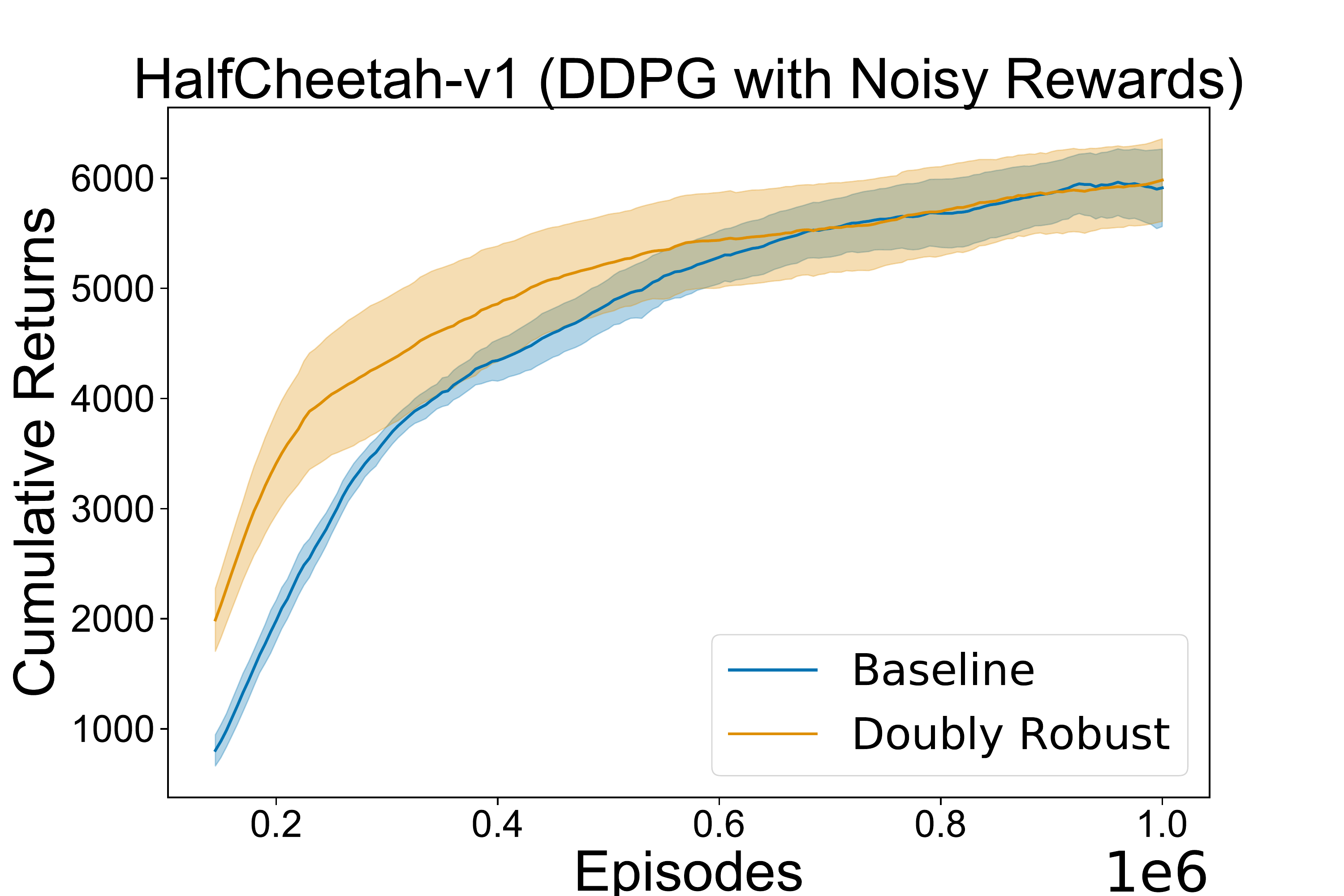}
    \!
   \includegraphics[width=0.3\linewidth]{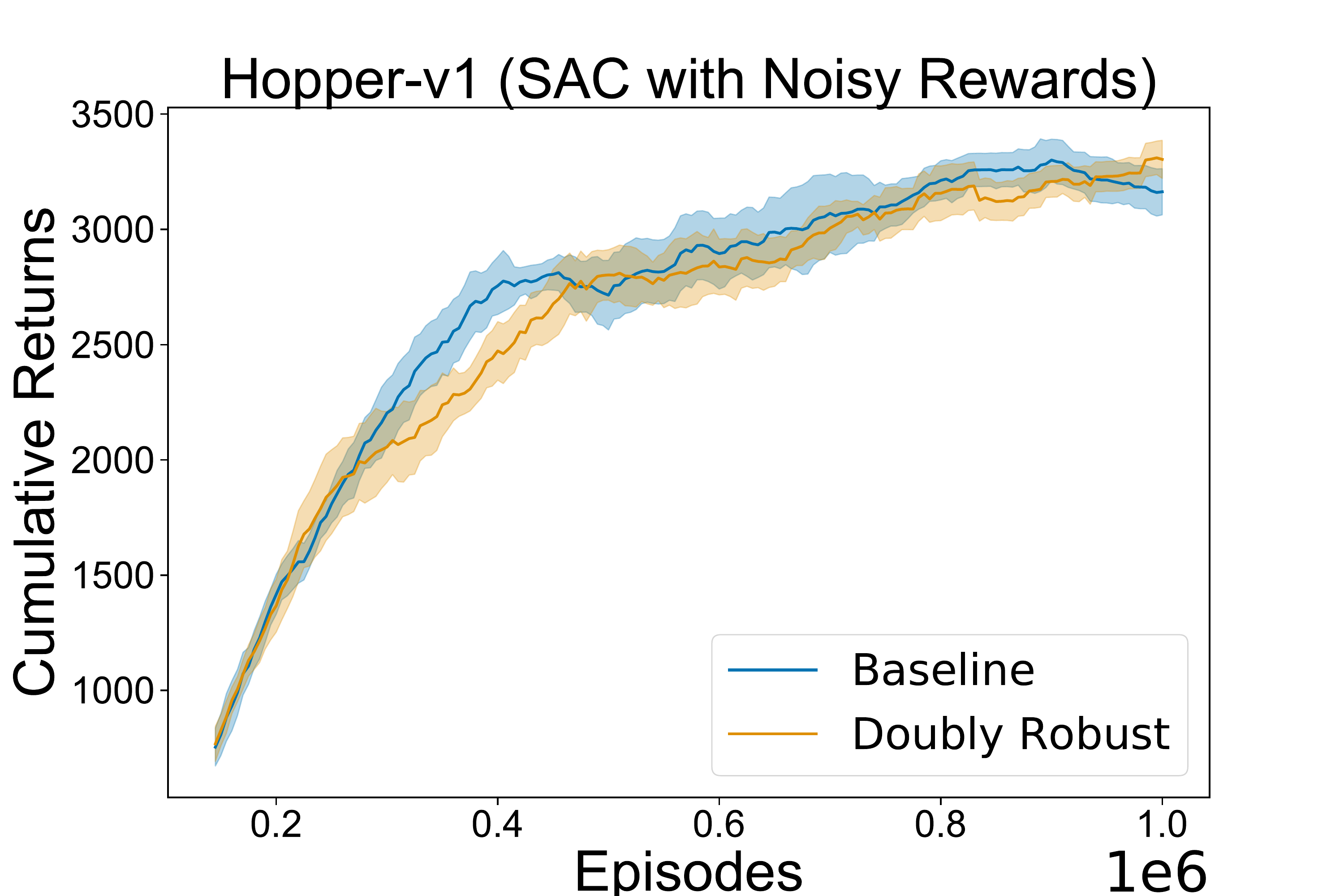}
   \!
   \includegraphics[width=0.3\linewidth]{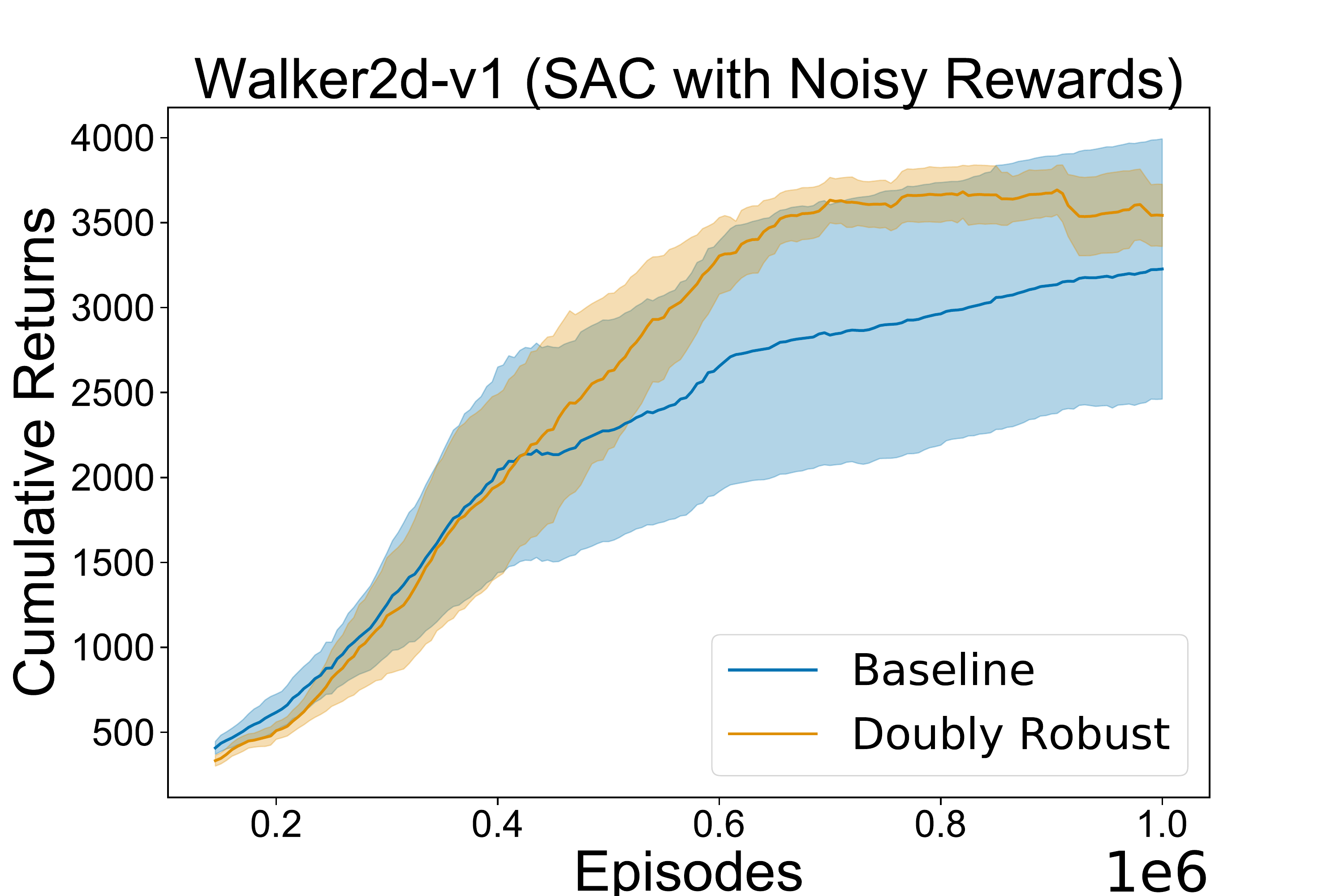}
   \!
   \includegraphics[width=0.3\linewidth]{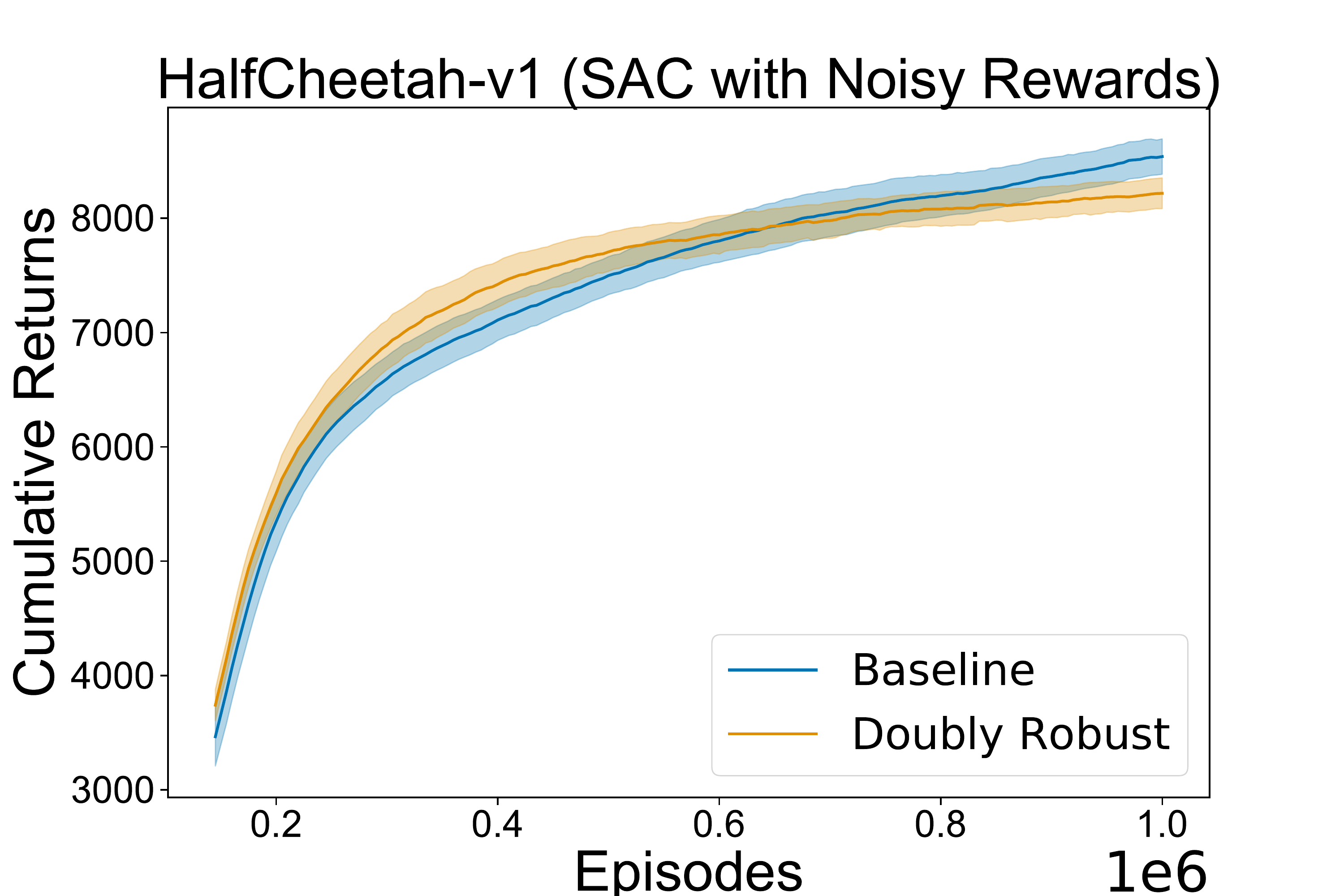}
   \!
   \includegraphics[width=0.3\linewidth]{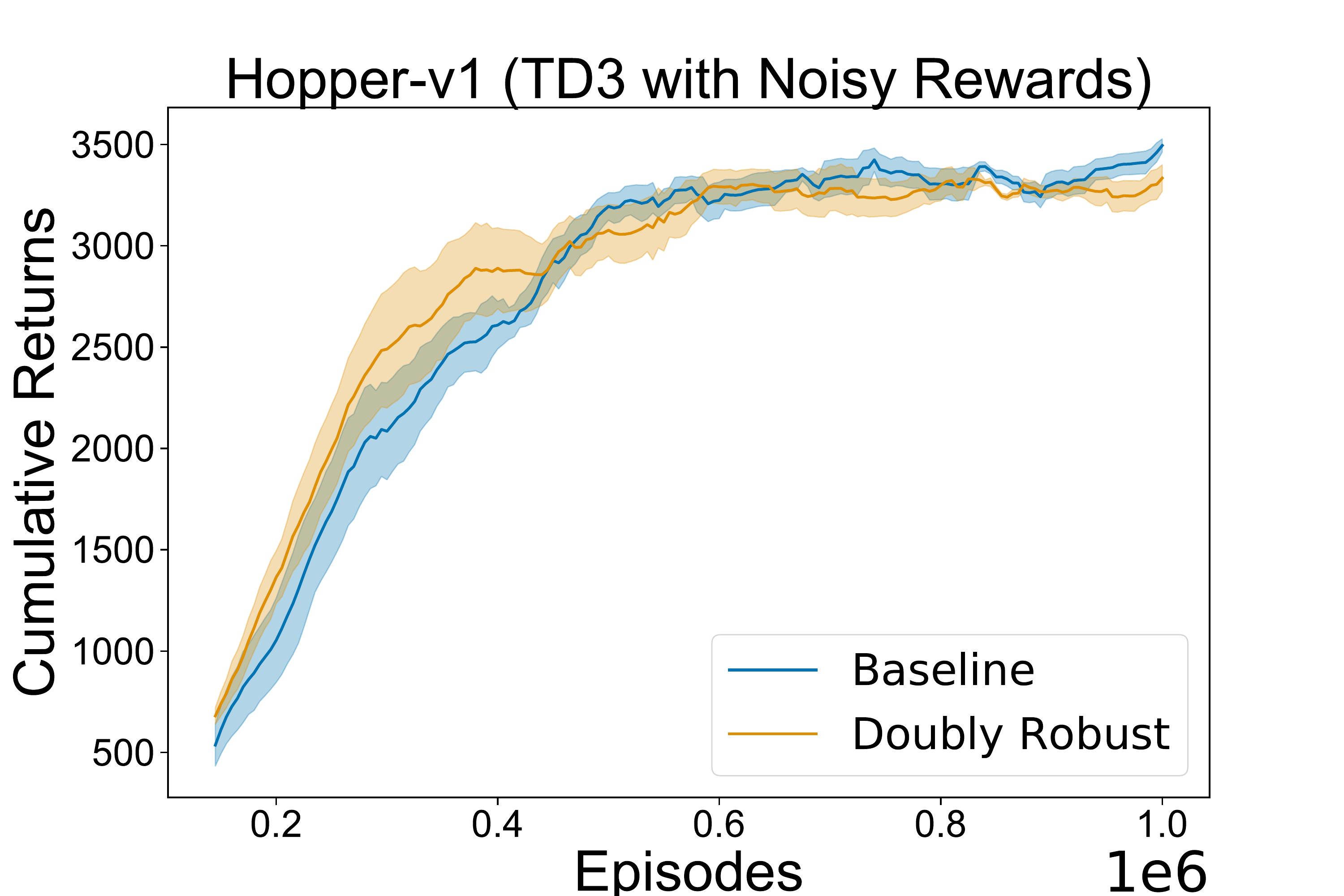}
     \!
   \includegraphics[width=0.3\linewidth]{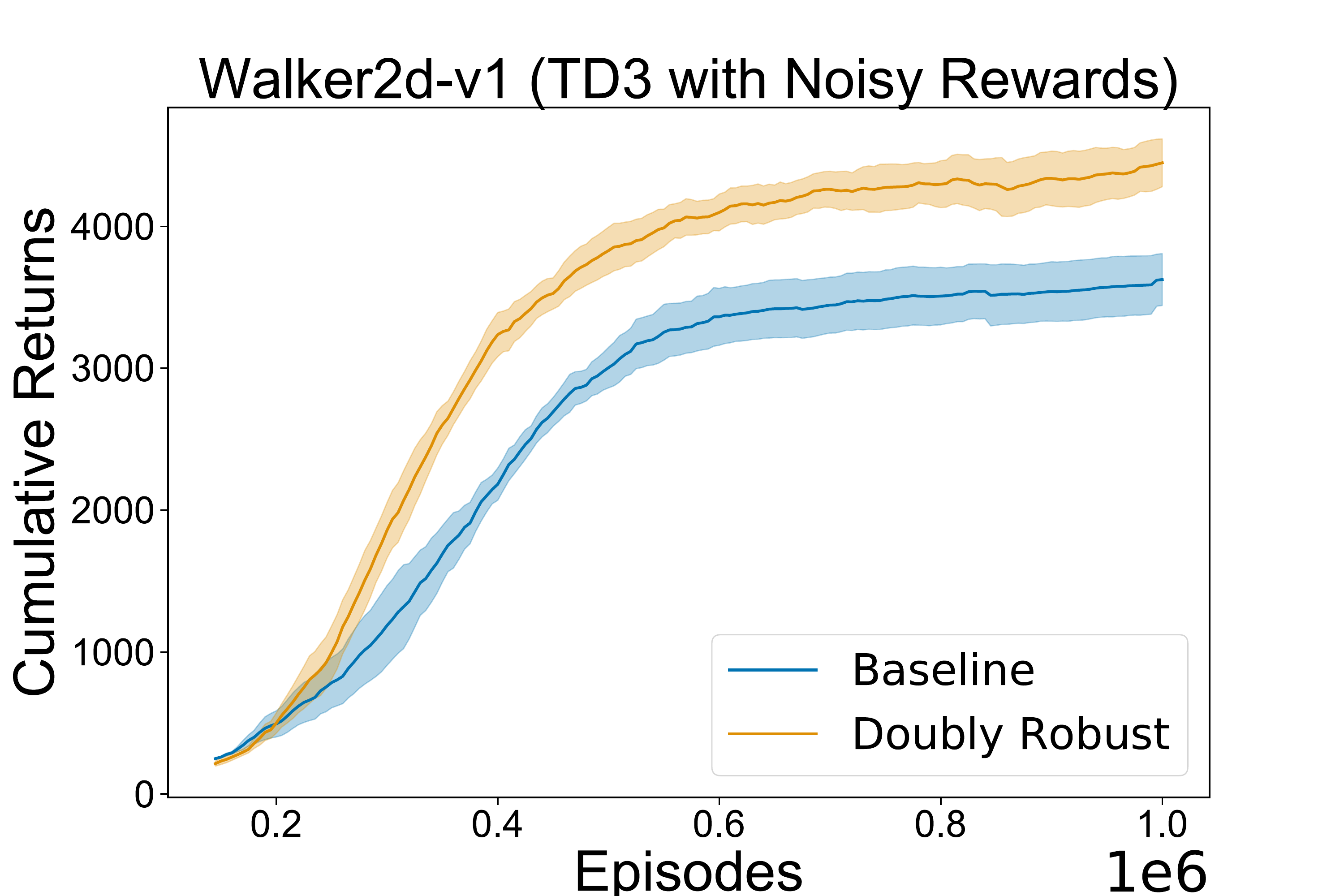}
   \!
    \includegraphics[width=0.3\linewidth]{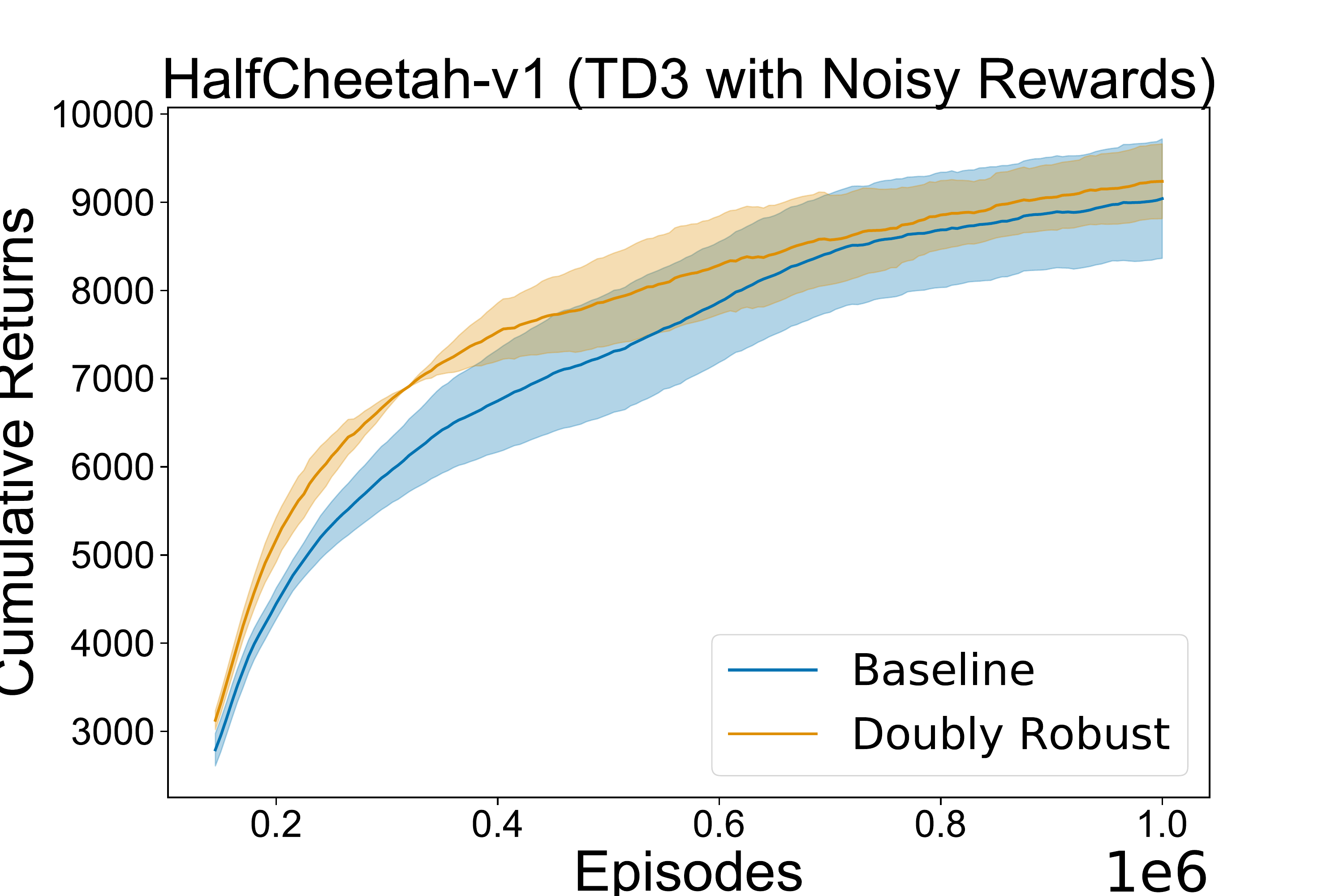}
\caption{Performance comparison of DDPG, SAC and TD3 with DR estimator using  noisy rewards, with $\sigma=1$}
\label{high_noisy_rewards}
\end{figure}

Figure \ref{low_noisy_rewards} and \ref{high_noisy_rewards} shows the performance comparison using DR estimator where reward prediction is done in the presence of a Gaussian noise added on top of true rewards. Since the reward estimator that we use performs best if the rewards are noisy we observe that the DR estimator thus obtained significantly outperforms in the majority of the mujoco tasks. We also run our DR estimator with different noise levels added to the rewards. In Figure \ref{low_noisy_rewards} shows the performance plots where the variance of the added Gaussian noise is $0.5$ and Figure \ref{high_noisy_rewards} shows the performance plots where the variance of the added Gaussian noise is $1.0$. We also observe that performance with DR estimator achieves lower variance compared to the baselines.

\section{Discussion and Conclusion}
\label{sec:conclusion}

We proposed an extension of doubly robust estimators from off-policy evaluation (OPE) methods to the off-policy policy gradient and actor-critic algorithms. This, we believe, is an important step towards extending the literature on OPE to the control setting. A large number of previous works on OPE are motivated for low variance but unbiased estimators of the value function. We highlight that such OPE estimators can also be extended to the actor-critic setting, where the critic evaluates the policy based on past data. 

Our proposed algorithm uses doubly robust estimators, for providing unbiased and low variance critic evaluation. This is particularly useful since majority of popular off-policy methods for control tasks rely on direct value based policy gradient estimates where having useful estimates of the value function plays an important role. We find that since the DR estimator plays the role of control variates to reduce variance of the critic estimate, it has a significant effect in terms of improving performance and lowering variance of existing popular off-policy gradient algorithms. We achieve DR estimation for the model-free setting by using a separate reward function estimator to predict the MDP rewards, since DR estimators use a combination of model-fre and model-based approaches. Our approach of estimating the rewards with a separate function approximator plays a further important role in settings where the reward function is stochastic and corrupted. We find that existing policy gradient algorithms can perform poorly in stochastic reward environments, due to high variance in the critic estimates. In such cases, DR estimators can be quite useful for further reducing the variance. Our algorithm plays an important step towards robust and safe RL methods, which is a crucial step for extending current advances of deep RL algorithms for real world applications.

For future work, it would be interesting to see other ways the reward function approximator can be used to predict the rewards $\hat{R}$. Since predicting the rewards can be considered as a supervised learning problem, it would be interesting to see the effect of over-fitting in the reward function approximation. Furthermore, we would require theoretical analysis of the bias variance trade-off of the DR estimator in the actor update, to fully understand the potential of DR estimators in the actor-critic setting, similar to previous studies of DR and variants of DR in terms of bias-variance typically done in OPE methods.

\bibliography{main}
\bibliographystyle{unsrtnat}

\end{document}